\documentclass{article}
\usepackage{microtype}
\usepackage{graphicx}
\usepackage{subfig}
\usepackage{booktabs} % for professional tables
\usepackage{hyperref}
\usepackage{url}
\usepackage{graphicx} % DO NOT CHANGE THIS
\usepackage{multirow}
\usepackage{amsmath}
\usepackage[utf8]{inputenc} % allow utf-8 input
\usepackage[T1]{fontenc}    % use 8-bit T1 fonts
\usepackage{nameref}
\usepackage{caption} % DO NOT CHANGE THIS AND DO NOT ADD ANY OPTIONS TO IT
\usepackage{varioref}
\usepackage{hyperref}
\usepackage{cleveref}
\usepackage{url}            % simple URL typesetting
\usepackage{booktabs}       % professional-quality tables
\usepackage{amsfonts}       % blackboard math symbols
\usepackage{nicefrac}       % compact symbols for 1/2, etc.
\usepackage{microtype}      % microtypography
\usepackage{lipsum}
\usepackage{float}
\usepackage{amssymb}% http://ctan.org/pkg/amssymb
\usepackage{pifont}% http://ctan.org/pkg/pifont
\newcommand{\cmark}{\ding{51}}%
\newcommand{\xmark}{\ding{55}}%

\usepackage{natbib}
\setcitestyle{authoryear,open={(},close={)}}
\usepackage{appendix}
\usepackage{multirow, booktabs}
\usepackage[table]{xcolor}
\newcommand{\RomanNumeralCaps}[1]
    {\MakeUppercase{\romannumeral #1}}
\newcommand\numberthis{\addtocounter{equation}{1}\tag{\theequation}}    
\newlength{\tempheight}
\newlength{\tempwidth}
\usepackage{amsmath}
\DeclareMathOperator*{\argmax}{arg\,max}

\newcommand{\rowname}[1]% #1 = text
{\rotatebox{90}{\makebox[\tempheight][c]{\textbf{#1}}}}

\newcommand{\columnname}[1]% #1 = text
{\makebox[\tempwidth][c]{\textbf{#1}}}

\usepackage{hyperref}

\usepackage[accepted]{icml2021}

\icmltitlerunning{CDT: Cascading Decision Trees for Explainable Reinforcement Learning}

\begin{document}

\twocolumn[
\icmltitle{CDT: Cascading Decision Trees for Explainable Reinforcement Learning}



\icmlsetsymbol{equal}{*}

\begin{icmlauthorlist}
\icmlauthor{Zihan Ding}{work}
\icmlauthor{Pablo Hernandez-Leal}{bor}
\icmlauthor{Gavin Weiguang Ding}{bor}
\icmlauthor{Changjian Li}{bor}
\icmlauthor{Ruitong Huang}{bor}
% \icmlauthor{Tateu H.~Yasehe}{ed,to,goo}
% \icmlauthor{Aaoeu Iasoh}{goo}
% \icmlauthor{Buiui Eueu}{ed}
% \icmlauthor{Aeuia Zzzz}{ed}
% \icmlauthor{Bieea C.~Yyyy}{to,goo}
% \icmlauthor{Teoau Xxxx}{ed}
% \icmlauthor{Eee Pppp}{ed}
\end{icmlauthorlist}

\icmlaffiliation{work}{Work done as an intern at Borealis AI.}
\icmlaffiliation{bor}{Borealis AI, Toronto, Canada}

\icmlcorrespondingauthor{Zihan Ding}{zhding@mail.ustc.edu.cn}
% \icmlcorrespondingauthor{Eee Pppp}{ep@eden.co.uk}

\icmlkeywords{Machine Learning, ICML}

\vskip 0.3in
]



\printAffiliationsAndNotice{\icmlEqualContribution} % otherwise use the standard text.

\begin{abstract}
Deep Reinforcement Learning (DRL) has recently achieved significant advances in various domains. However, explaining the policy of RL agents still remains an open problem due to several factors, one being the complexity of explaining neural networks decisions. Recently, a group of works have used decision-tree-based models to learn explainable policies. Soft decision trees (SDTs) and discretized differentiable decision trees (DDTs) have been demonstrated to achieve both good performance and share the benefit of having explainable policies. In this work, we further improve the results for tree-based explainable RL in both performance and explainability. Our proposal, Cascading Decision Trees (CDTs) apply representation learning on the decision path to allow richer expressivity. Empirical results show that in both situations, where CDTs are used as policy function approximators or as imitation learners to explain black-box policies, CDTs can achieve better performances with more succinct and explainable models than SDTs. As a second contribution our study reveals limitations of explaining black-box policies via imitation learning with tree-based explainable models, due to its inherent instability.
\end{abstract}

\section{Introduction}
\label{intro}
%Deep neural networks (DNN) powered algorithms have demonstrated its effectiveness and become pervasive across different domains, including vision recognition, credit assessment, and even diseases diagnostics. However, along with its great modeling capacity, the use of DNNs makes the users have no access to a satisfactory explanation of the model.
%When the models are deployed in the domains where the accountability of the decisions is critical, for example in healthcare or in law enforcement, explanation for the decisions made by the system may outweighs the precision so that trust can be built between the users and the model. The demand of such explanability in developing machine learning algorithms has recently attracted lots of attention, and a new research topic, Explainable Artificial Intelligence (XAI), has emerged in the literature. 

%Currently, XAI research has been mostly focusing on the explanability of complicated models in supervised learning. Especially, [image classification, NLP].  On the other hand, explainable reinforcement learning (XRL) is relatively less explored. How to interpret the action choices in reinforcement learning (RL) policies remains a critical challenge, especially as the gradually increasing trend of applying RL in various domains involving transparency and safety. ~\cite{puiutta2020explainable} [Why it is different from that in supervised learning?] \citep{samek2019explainable}

Explainable Artificial Intelligence (XAI), especially Explainable Reinforcement Learning (XRL)~\citep{puiutta2020explainable} is attracting more attention recently.
How to interpret the action choices in reinforcement learning (RL) policies remains a critical challenge, especially as the gradually increasing trend of applying RL in various domains involving transparency and safety~\citep{cheng2019end, junges2016safety}.
Currently, many state-of-the-art DRL agents use neural networks (NNs) as their function approximators. While NNs are considered stronger function approximators (for better performances), RL agents built on top of them are generally lack of interpretability~\citep{lipton2018mythos}. Indeed, interpreting the behavior of NNs themselves remains an open problem in the field~\citep{montavon2018methods, albawi2017understanding}. 

% The Decision Tree (DT) is one typical type of model that helps human to interpret the decision making process via visualizing the decision path. 
In contrast, traditional DTs (with hard decision boundaries) are usually regarded as models with readable interpretations for humans, since humans can interpret the decision making process by visualizing the decision path. However, DTs may suffer from weak expressivity and therefore low accuracy.  
% Differentiable DTs~\citep{frosst2017distilling} lie in the middle of the two and have gained increased interest, especially for their apparent improved interpretability. \cite{suarez1999globally} first propose soft/fuzzy DT (shorten as SDT), as the most primitive paper.
An early approach to reduce the hardness of DT was the soft/fuzzy DT (shorten as SDT) proposed by \cite{suarez1999globally}. Recently, differentiable SDTs~\citep{frosst2017distilling} have shown both improved interpretability and better function approximation, which lie in the middle of traditional DTs and neural networks.

People have adopted differentiable DTs for interpreting RL policies in two slightly different settings: an imitation learning setting~\citep{coppens2019distilling, liu2018toward}, in which imitators with interpretable models are learned from RL agents with black-box models, or a full RL setting~\citep{silva2019optimization}, where the policy is directly represented as an interpretable model, \emph{e.g.}, DT. However, the DTs in these methods only conduct partitions in raw feature spaces without representation learning that could lead to complicated combinations of partitions, possibly hindering both model interpretability and scalability. Even worse, some methods have axis-aligned partitions (univariate decision nodes)~\citep{wu2017beyond, silva2019optimization} with much lower model expressivity.

% We propose Cascading Decision Trees (CDTs) strikes a balance between the model interpretability and accuracy, with adequate representation learning also based on interpretable models (\emph{e.g.} linear models). It is testified to have the benefits of a significantly smaller number of parameters and a more compact tree structure with even better performance. 
In this paper, we propose Cascading Decision Trees (CDTs) striking a balance between model interpretability and accuracy, this is, having an adequate representation learning  based on interpretable models (\emph{e.g.} linear models). Our experiments show that CDTs share the benefits of having a significantly smaller number of parameters (and a more compact tree structure) and better performance than related works. The experiments are conducted on RL tasks, in either imitation-learning or RL settings. We also demonstrate that the imitation-learning approach is less reliable for interpreting the RL policies with DTs, since the imitating DTs may be prominently different in several runs, which also leads to divergent feature importances and tree structures.

\section{Related Works}

 A series of works were developed in the past two decades along the direction of differentiable DTs~\citep{irsoy2012soft, laptev2014convolutional}.
Recently, \citet{frosst2017distilling} proposed to distill a SDT from a neural network. Their approach was only tested on MNIST digit classification tasks. \citet{wu2017beyond} further proposed the tree regularization technique to favor the models with decision boundaries closer to compact DTs for achieving interpretability. To further boost the prediction accuracy of tree-based models, two main extensions based on single SDT were proposed: (1) ensemble of trees, or (2) unification of NNs and DTs. 

An ensemble of decision trees is a common technique used for increasing accuracy or robustness of prediction, which can be incorporated in SDTs~\citep{rota2014neural, kontschieder2015deep, kumar2016ensemble}, giving rise to neural decision forests. Since more than one tree needs to be considered during the inference process, this might yield complications in the interpretability. A common solution is to transform the decision forests into a single tree~\citep{sagi2020explainable}.

As for the unification of NNs and DTs, \citet{laptev2014convolutional} propose convolutional decision trees for feature learning from images. Adaptive Neural Trees (ANTs)~\citep{tanno2018adaptive} incorporate representation learning in decision nodes of a differentiable tree with nonlinear transformations like convolutional neural networks (CNNs). The nonlinear transformations of an ANT, not only in routing functions on its decision nodes but also in feature spaces, guarantee the prediction performances in classification tasks on the one hand, but also hinder the potential of interpretability of such methods on the other hand. \cite{wan2020nbdt} propose the neural-backed decision tree (NBDT) which transfers the final fully connected layer of a NN into a DT with induced hierarchies for the ease of interpretation, but shares the convolutional backbones with normal deep NNs, yielding the state-of-the-art performances on CIFAR10 and ImageNet classification tasks.

However, these advanced methods either employ multiple trees with multiplicative numbers of model parameters, or heavily incorporate deep learning models like CNNs in the DTs. Their interpretability is severely hindered due to their model complexity. %Therefore these models are not good candidates for interpreting the RL policies. 

To interpret an RL agent, \citet{coppens2019distilling} propose distilling the RL policy into a differentiable DT by imitating a pre-trained policy. Similarly, \citet{liu2018toward} apply an imitation learning framework but to the $Q$ value function of the RL agent. They also propose Linear Model U-trees (LMUTs) which allow linear models in leaf nodes. 
% whose weights are updated with Stochastic Gradient Descent (the rest of the nodes follow the same criteria as in traditional trees). 
\citet{silva2019optimization} propose to apply differentiable DTs directly as function approximators for either $Q$ function or the policy in RL. They apply a discretization process and a rule list tree structure to simplify the trees for improving interpretability. The VIPER method proposed by \cite{bastani2018verifiable} also distills policy as NNs into a DT policy with theoretically verifiable capability, but for imitation learning settings and nonparametric DTs only.

% Recently, \citet{ellis2020dreamcoder} proposed DreamCoder, which learns to solve general problems with interpretability through writing programs. The final program for a task is composed of hierarchically organized layers of concepts, whereas those concepts are conceptual abstractions that represent common fragments across task solutions, as well as the primitives for the programming languages. The hierarchical conceptual abstraction and the program solutions built on top of that also make the DreamCoder a well-interpreted method.

% Compared with three main categories of methods with differentiable DTs for XRL, the distinctiveness of our proposed CDT is as follows:
% The key uniqueness in the idea of cascading differential decision tree is listed below, compared with three main baselines with each representing a category of methods with different characteristics on differentiable decision tree:
Our proposed CDT is distinguished from other main categories of methods with differentiable DTs for XRL in the following ways:
(i) Compared with SDT~\citep{frosst2017distilling}, partitions in CDT not only happen in original input space, but also in transformed spaces by leveraging intermediate features. This is well documented in recent works~\citep{kontschieder2015deep, xiao2017ndt, tanno2018adaptive} to improve model capacity, and it can be further extended into hierarchical representation learning with advanced feature learning modules like CNN~\citep{tanno2018adaptive}. (ii) Compared with work by \citet{coppens2019distilling}, space partitions are not limited to axis-aligned ones (which hinders the expressivity of trees with certain depths), but achieved with linear models of features as the routing functions. Moreover, the adopted linear models are not a restriction (but as an example) and other interpretable transformations are also allowed in our CDT method. (iii) Compared with ANTs~\citep{tanno2018adaptive}, our CDT method unifies the decision making process based on different intermediate features with a single decision making tree, which follows the low-rank decomposition of a large matrix with linear models. It thus greatly improves the model simplicity for achieving interpretability. 
% About model simplicity and interpretability in DTs, see our motivating example in Appendix A.

% the duplicative structure in decision making process based on different intermediate features is summarized into a single decision making tree, rather than allowing fully relaxed freedoms in tree structures as end-to-end solutions from raw inputs to final outputs (which hinders interpretability due to the model complexity). Although the duplicative structure may not always exist in all tasks, it can be observed in the heuristic agents of some RL tasks. And this is the key to simplify the tree-based model for interpretability in a CDT. %{\bf Point (iii) is not very clear}

\section{Method}
% We give a brief introduction of SDT and introduce our method CDT in this section. 

% Before presenting CDTs in detail, we first show a motivating example on DTs and interpretability.

% Heuristic agent in \textit{LunarLander-v2} can be transformed into a multivariate decision tree, so we choose the environment as the testbed.

% \subsection{Policy Distillation}
% Policy distillation into a SDT (SDT) or soft decision forest (SDF), instead of another neural network.

% Both SDT and SDF are tried in our experiment. Although a deep SDT or a SDF as an ensemble of trees are normally viewed as methods that could be less interpretable, we develop a feature importance assignment method to fairly distribute the importance on feature dimensions for increasing interpretability.

% Several contents are involved in this work, summarized as follows:

% 1. We first need to clarify the simplicity and its relationship with interpretability of DT models, so as to form not only the preferences in designing the tree structures but also criteria for evaluating them.

% 2. We dive into the approach of SDT as imitating model of RL policies, to evaluate the reliability of deploying it for achieving interpretability.

% 3. We propose CDT as policy function approximators, and experimentally testify its advantages in model simplicity and performance.

\subsection{Soft Decision Tree (SDT)}
\begin{figure}[htbp]
    \begin{center}
        \includegraphics[scale=0.18]{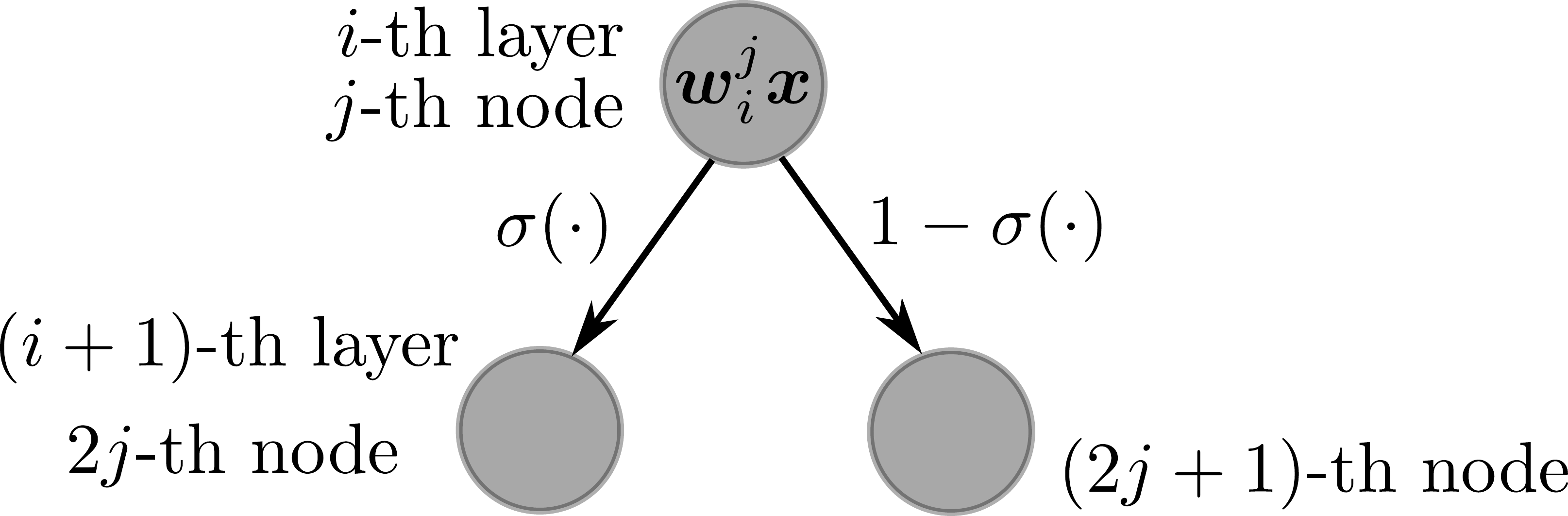}
    \end{center}
    \caption{A SDT node. $\sigma(\cdot)$ is the sigmoid function with function values on decision nodes as input.}
    \label{fig:sdt_node}
\end{figure}

% Advantages: the SDT can incorporate feature priors as decision boundaries, either linear or non-linear?
A SDT is a differentiable DT with a probabilistic decision boundary at each node. Considering we have a DT of depth $D$, each node in the SDT can be represented as a weight vector (with the bias as an additional dimension) $\boldsymbol{w}^j_i$, where $i$ and $j$ indicate the index of the layer and the index of the node in that layer respectively, as shown in Fig.~\ref{fig:sdt_node}. The corresponding node is represented as $n_u$, where $u=2^{i-1}+j$ uniquely indices the node. 

The decision path for a single instance can be represented as set of nodes $\mathcal{P}\subset\mathcal{N}$, where $\mathcal{N}$ is the set for all nodes on the tree. We have $\mathcal{P} = \argmax_{\{u\}} \prod_{i=1}^D p^{\lfloor j/2 \rfloor \rightarrow j}_{i-1\rightarrow i}$, where $p^{\lfloor j/2 \rfloor \rightarrow j}_{i-1\rightarrow i}$ is the probability of going from node $n_{2^{i-2}+\lfloor j/2 \rfloor}$ to $n_{2^{i-1}+j}$. The $\{u\}$ indicates that the $\arg\max$ is taken over a set of nodes rather than a single one. Note that $p^{\lfloor j/2 \rfloor \rightarrow j}_{i-1\rightarrow i}$ will always be 1 for a hard DT~\citep{safavian1991survey}. Therefore the path probability to a specific node $n_u$ is: $P^u=\prod_{i^\prime=1}^{j^\prime} p^{\lfloor j^\prime/2 \rfloor \rightarrow j^\prime}_{i^\prime-1\rightarrow i^\prime}, u^\prime\in \mathcal{P}$. In the following, we name all DTs using probabilistic decision path as SDT-based methods, shorten as SDT.

\citet{silva2019optimization} propose to discretize the learned differentiable SDTs into univariate DTs for improving interpretability. Specifically, for a decision node with a $(k+1)$-dimensional vector $\boldsymbol{w}$ (the first dimension $w_1$ is the bias term), the discretization process (i) selects the index of largest weight dimension as $k^*=\argmax_k w_k$ and (ii) divides $w_1$ by $w_{k^*}$, to construct a univariate hard DT. Without further description, the default discretization process in our experiments for both SDTs and CDTs also follows this manner. The SDTs are therefore the same as DDTs in \citet{silva2019optimization}.

\subsection{Cascading Decision Tree (CDT)}
\subsubsection{Motivating Examples}
Before introducing our method, we first show two simple examples to demonstrate our motivations for proposing CDT.

\textbf{Multivariate \emph{v.s.} Univariate Decision Nodes}
People have proposed a variety of desiderata for interpretability~\citep{lipton2018mythos}, including trust, causality, transferability, informativeness, etc. Here we summarize the answers in general into two aspects: (1) interpretable meta-variables that can be directly understood; (2) model simplicity. Human-understandable variables with simple model structures comprise most of the models interpreted by humans either in a form of physical and mathematical principles or human intuitions, which is also in accordance with the Occam's razor principle.
\begin{figure}[!htbp]
    \begin{center}
        \includegraphics[scale=1.2]{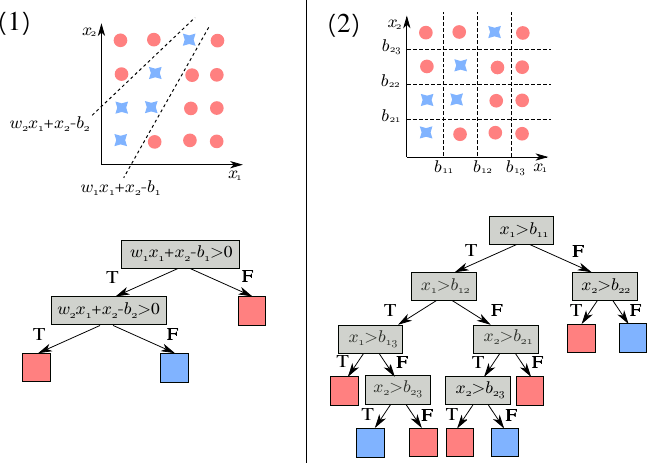}
    \end{center}
    \caption{Comparison of two different tree structures on a binary classification problem.}
    \label{fig:2trees}
\end{figure}
For model simplicity, a simple model in most cases is more interpretable than a complicated one. Different metrics can be applied to measure the model complexity~\citep{murray2007reducing, molnar2019quantifying}, like the number of model parameters, model capacity, computational complexity, non-linearity, etc. There are ways to reduce the model complexity: model projection from a large space into a small sub-space, model distillation, merging the replicates in the model, etc. Feature importance~\citep{schwab2019cxplain} (\emph{e.g.}, through estimating the sensitivity of model outputs with respect to inputs) is one type of methods for projecting a complicated high-dimensional parameter space into a scalar space across feature dimensions. The proposed method CDT in this paper is a way to improve model simplicity by merging the replicates through representation learning.

To clarify our choice of model in terms of simplicity, we show an example in Fig.~\ref{fig:2trees} to compare different types of decision trees involving univariate and multivariate decision nodes. It shows the comparison of a multivariate DT and a univariate DT for a binary classification task. Apparently, the multivariate DT is simpler than the univariate one in its structure, as well as with fewer parameters, which makes it potentially more interpretable. For even more complex cases, the multivariate tree structure is more likely to achieve necessary space partitioning with simpler model structures. 

\begin{figure}[htbp]
    \begin{center}
        \includegraphics[scale=0.18]{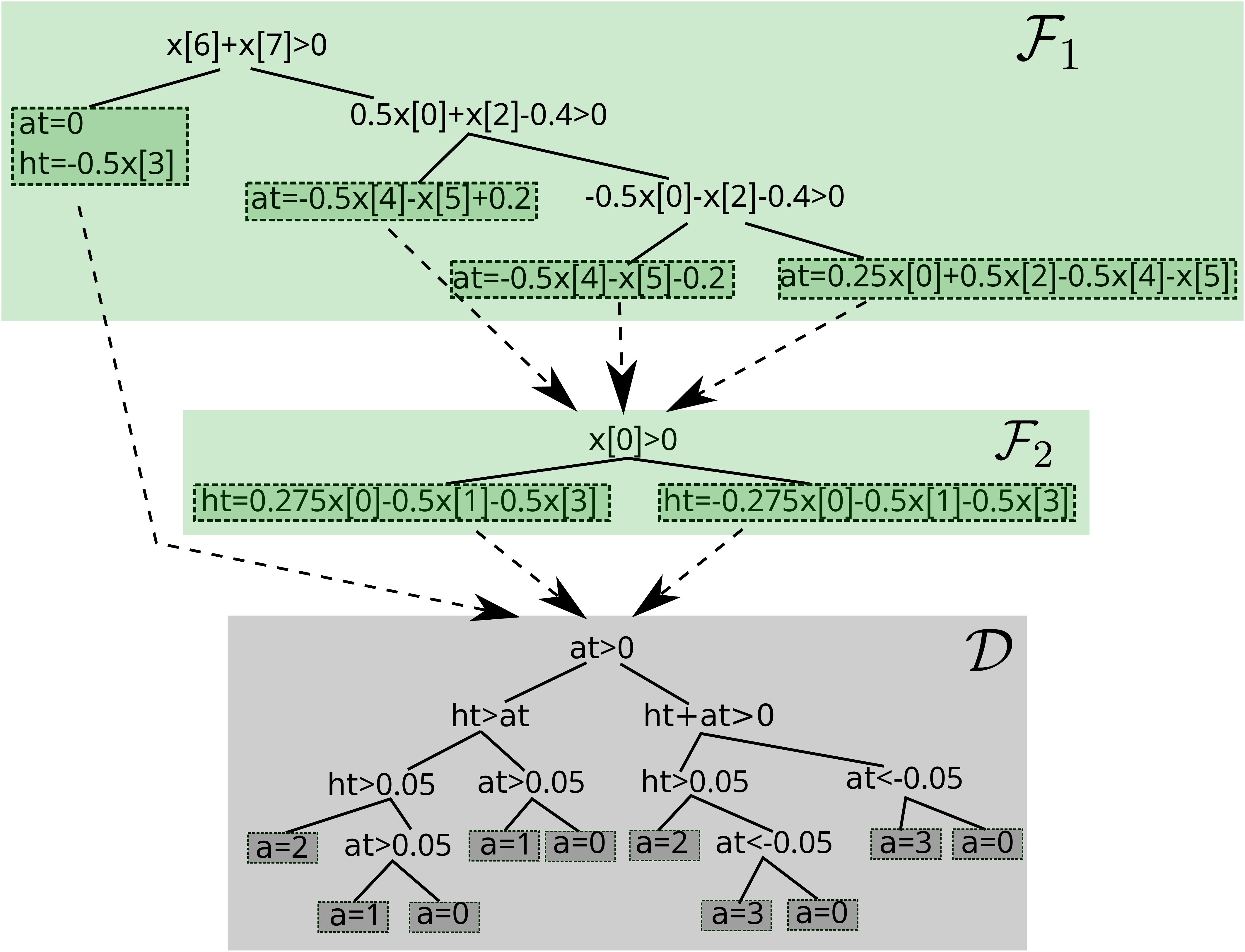}
    \end{center}
    \caption{The heuristic decision tree for \textit{LunarLander-v2}, with the left branch and right branch of each node yielding the "True" and "False" cases respectively. $x$ is an 8-dimensional observation, $a$ is the univariate discrete action given by the agent. $at, ht$ are two intermediate variables, corresponding to the "angle-to-do" and "hover-to-do" in the heuristic solution.}
    \label{fig:lunarlander_hdt}
\end{figure}

\textbf{Intermediate Feature}
As shown in Fig.~\ref{fig:lunarlander_hdt}, we analyzed the heuristic solution\footnote{In the code repository of OpenAI Gym: https://github.com/openai/gym/blob/master/gym/envs/box2d/lunar\newline\_lander.py} of \textit{LunarLander-v2} and found that it contains duplicative structures after being transformed into a decision tree, which can be leveraged to simplify the models to be learned. Specifically, the two green modules $\mathcal{F}_1$ and $\mathcal{F}_2$ in the tree are basically assigning different values to two intermediate variables ($ht$ and $at$) under different cases, while the grey module $\mathcal{D}$ takes the intermediate variables to achieve action selection. The modules $\mathcal{F}_2$ and $\mathcal{D}$ are used repeatedly on different branches on the tree, which forms a duplicative structure. This can help with the simplicity and interpretability of the model, which motivates our idea of CDT methods for XRL.

\subsubsection{Methods}
We propose CDT as an extension based on SDT with multivariate decision nodes, allowing it to have the capability of representation learning as well as decision making in transformed spaces. In a simple CDT architecture as shown on the left of Fig.~\ref{fig:cascade_merge}, a feature learning tree $\mathcal{F}$ is cascaded with a decision making tree $\mathcal{D}$. 
% Here we represent SDT in a more general way, and CDT is an extension based on SDT with two trees in a cascading structure: a feature learning tree $\mathcal{F}$ and a decision making tree $\mathcal{D}$, as shown in Fig~\ref{fig:cascade_merge}. 
In tree $\mathcal{F}$, each decision node is a simple function of raw feature vector $\boldsymbol{x}$ given learnable parameters $\boldsymbol{w}$: $\phi(\boldsymbol{x};\boldsymbol{w})$, while each leaf of it is a feature representation function: $\boldsymbol{f}=f(\boldsymbol{x};\tilde{\boldsymbol{w}})$ parameterized by $\tilde{\boldsymbol{w}}$. In tree $\mathcal{D}$, each decision node is a simple function of learned features $\boldsymbol{f}$ rather than raw features $\boldsymbol{x}$ given learnable parameters $\boldsymbol{w}^\prime$: $\psi(\boldsymbol{f};\boldsymbol{w}^\prime)$. The output distribution of $\mathcal{D}$ is another parameterized function $p(\cdot;\tilde{\boldsymbol{w}}^\prime)$ independent of either $\boldsymbol{x}$ or $\boldsymbol{f}$. For simplicity and interpretability, all functions $\phi, f \text{ and } \psi$ are linear functions in our examples, but they are free to be extended with other interpretable models.

Specifically, we provide detailed mathematical relationships based on linear functions as follows. For an environment with input state vector $\boldsymbol{x}$ and output discrete action dimension $O$, suppose that our CDT has intermediate features of dimension $K$ (not the number of leaf nodes on $\mathcal{F}$, but for each leaf node), we have the probability of going to the left/right path on the $u$-th node on $\mathcal{F}$: 
% The learned intermediate features $\{f_k| k=0, 1, ..., K-1\}$ are linear combinations (no bias terms) of the raw input features $\{x_r| r=0, 1, ..., R-1\}$ with $R$ dimensions as:

\begin{align}
    p^{\text{Go Left}}_u = \sigma({\boldsymbol{w}_{k}\cdot \boldsymbol{x}}), \quad p^{\text{Go Right}}_u = 1-p^{\text{Go Left}}_u,
    \label{eq:1}
\end{align}
% \begin{align}
%     p^{\text{Go Right}}_u = 1-p^{\text{Go Left}}_u, 
% \end{align}

which is the same as in SDTs. Then we have the linear feature representation function for each leaf node on $\mathcal{F}$, which transforms the basis of the representation space with:
\begin{figure}[h!]
    \begin{center}
        \includegraphics[scale=0.4]{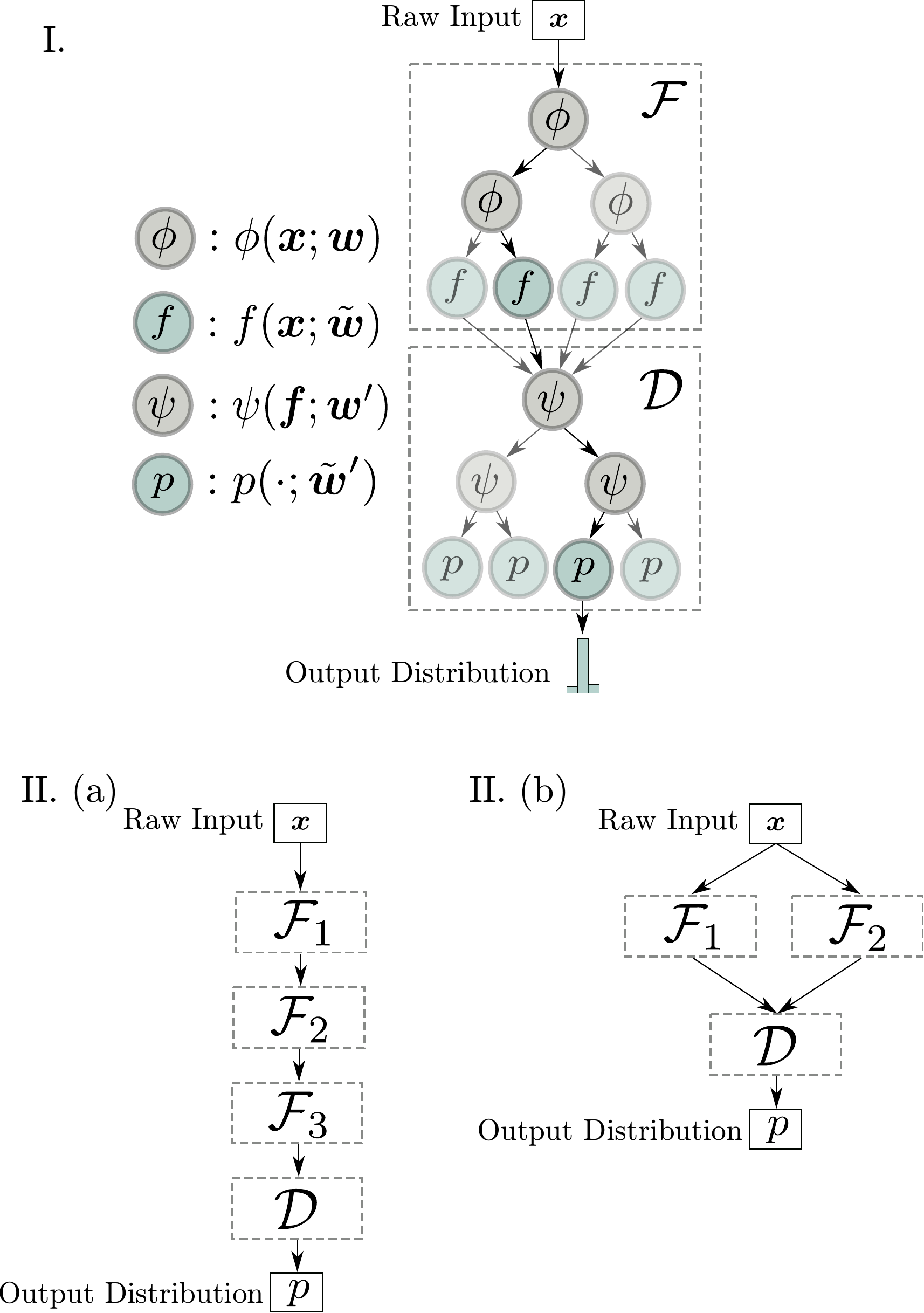}
    \end{center}
    \caption{CDT methods. I. is a simple CDT architecture, consisting of a feature learning tree $\mathcal{F}$ and a decision making tree $\mathcal{D}$; II. shows two possible types of hierarchical CDT architectures, where (a) is an example architecture with hierarchical representation learning using three cascading $\mathcal{F}$ before one $\mathcal{D}$, and (b) is an example architecture with two $\mathcal{F}$ in parallel, potentially with different dimensions of $x$ as inputs.}
    \label{fig:cascade_merge}
\end{figure}
\begin{align}
    f_k = \tilde{\boldsymbol{w}}_{k}\cdot \boldsymbol{x},  k=0,1, ..., K-1
    \label{eq:2}
\end{align}

which gives the $K$-dimensional intermediate feature vector $\boldsymbol{f}$ for each possible path. Due to the symmetry in all internal layers within a tree, all internal nodes satisfy the formulas in Eq.~(\ref{eq:1})(\ref{eq:2}). In tree $\mathcal{D}$, it is also a SDT but with raw input $\boldsymbol{x}$ replaced by learned representations $\boldsymbol{f}$ for each node $u^\prime$ in $\mathcal{D}$:

\begin{align}
    p^{\text{Go Left}}_{u^\prime} = \sigma(\tilde{\boldsymbol{w}}_{k}\cdot \boldsymbol{f}), \quad p^{\text{Go Right}}_{u^\prime} = 1-p^{\text{Go Left}}_{u^\prime}, 
    \label{eq:3}
\end{align}
% \begin{align}
%     p^{\text{Go Right}}_{u^\prime} = 1-p^{\text{Go Left}}_{u^\prime}, 
% \end{align}

Finally, the output distribution is feature-independent, which gives the probability mass values across output dimension $O$ for each leaf of $\mathcal{D}$ as: 
\begin{align}
    p_{k^\prime} = \frac{\exp (\tilde{\boldsymbol{w}}^\prime)}{\sum_{k^\prime=0}^{O-1} \exp (\tilde{\boldsymbol{w}}^\prime_{k^\prime})}, k^\prime=0,1,...,O-1
\end{align}

% \begin{align}
%     f_k = \sum_{r=0, 1, ..., R-1} \tilde{w}_{kr}x_r
% \end{align}
% So each leaf of the feature learning tree assigns the intermediate features with different values via different weights. For example, for the $l$-th leaf, the feature set will be $\{f^l_k| k=0, 1, ..., K-1\}$ with leaf weights $\{\tilde{w}^l_{kr}|k=0, 1, ..., K-1, r=0, 1, ..., R-1\}$.

% In the decision making trees, those learned intermediate features will be taken as normal input features to a differentiable DT. During training, we use the expected intermediate features from all leaves in the feature learning tree as input features to the decision making tree:
% \begin{align}
%     f_k &= \phi(\boldsymbol{x}) \\
%     &= \sum_{l=0}^{L-1} P^lf^l_k, k=0,1,...,K-1
% \end{align}
% where $P_l$ is the path probability from the root node to the $l$-th leaf node in the feature learning tree, and $f^l_k$ is the $k$-th dimension of intermediate feature for the $l$-th leaf in the feature learning tree. 

Suppose we have a CDT of depth $N_1$ for $\mathcal{F}$ and depth $N_2$ for $\mathcal{D}$, the probability of going from root of either $\mathcal{F}$ or $\mathcal{D}$ to $u$-th leaf node on each sub-tree both satisfies previous derivation in SDTs:
 $P^u=\prod_{i^\prime=1}^{j^\prime} p^{\lfloor j^\prime/2 \rfloor \rightarrow j^\prime}_{i^\prime-1\rightarrow i^\prime}, u^\prime\in \mathcal{P}$, where $\mathcal{P}$ is the set of nodes on path. Therefore the overall path probability of starting from the root of $\mathcal{F}$ to $u_1$-th leaf node of $\mathcal{F}$ and then $u_2$-th leaf node of $\mathcal{D}$ is:
 \begin{align}
    P = P^{u_1}P^{u_2}
\end{align}

Each leaf of the feature learning tree represents one possible assignment for intermediate feature values, while they share the subsequent decision making tree.
During the inference process, we simply take the leaf on $\mathcal{F}$ or $\mathcal{D}$ with the largest probability to assign values for intermediate features (in $\mathcal{F}$) or derive output probability (in $\mathcal{D}$), which may sacrifice little accuracy but increase interpretability. The detailed architecture of CDT with relationships among variables is plotted in figures in Appendix A.

\textbf{Model Simplicity}

We analyze the simplicity of CDT compared with SDT in terms of the numbers of learnable parameters in the model. The reason for doing this is that in order to increase the interpretability, we need to simplify the tree structure or reduce the number of parameters including weights and bias in the tree. 

We can analyze the model simplicity of CDT against a normal SDT with linear functions in a matrix decomposition perspective. Suppose we need a total of $M$ multivariate decision nodes in the $R$-dimensional raw input space $\mathbb{X}$ to successfully partition the space for high-performance prediction, which can be written as a matrix $\mathbf{W}^x_{M\times R}$. CDT tries to achieve the same partitions through learning a transformation matrix $\mathbf{T}_{K\times R}: \mathbb{X}\rightarrow \mathbb{F}$ for all leaf nodes in $\mathcal{F}$ and a partition matrix $\mathbf{W}^f_{M\times K}$ for all internal nodes in $\mathcal{D}$ in the $K$-dimensional feature space $\mathbb{F}$, such that:
\begin{align}
    \mathbf{W}^x \boldsymbol{x} &= \mathbf{W}^f \boldsymbol{f} =\mathbf{W}^f \mathbf{T} \boldsymbol{x}\\
    \Rightarrow \mathbf{W}^x & =\mathbf{W}^f \mathbf{T}
    \label{equ:matrix_decompos}
\end{align}
Therefore the number of model parameters to be learned with CDT is reduced by $M\times R - (M\times K + K \times R)$ compared against a standard SDT of the same total depth, and it is a positive value as long as $K<\frac{M\times R}{M+R}$, while keeping the model expressivity.

A detailed quantitative analysis of model parameters for CDT and SDT is provided in Appendix B.

\textbf{Hierarchical CDT}

From above, a simple CDT architecture as in Fig.~\ref{fig:cascade_merge} with a single feature learning model $\mathcal{F}$ and single decision making model $\mathcal{D}$ can achieve intermediate feature learning with a significant reduction in model complexity compared with traditional SDT. However, sometimes the intermediate features learned with $\mathcal{F}$ may be unsatisfying for capturing complex structures in advanced tasks, therefore we further extend the simple CDT architecture into more hierarchical ones. As shown on the right side in Fig.~\ref{fig:cascade_merge}, two potential types of hierarchical CDT are displayed: (a) a hierarchical feature abstraction module with three feature learning models $\{\mathcal{F}_1, \mathcal{F}_2, \mathcal{F}_3\}$ in a cascading manner before inputting to the decision module $\mathcal{D}$; (b) a parallel feature extraction module with two feature learning models $\{\mathcal{F}_1, \mathcal{F}_2\}$ before concatenating all learned features into $\mathcal{D}$. 

One needs to bear in mind that whenever the model structures are complicating, the interpretability of the model decreases due to the loss of simplicity. Therefore we did not apply the hierarchical CDTs in our experiments for maintaining interpretability. However, the hierarchical structure is one of the most preferred ways to keep simplicity as much as possible if trying to increase the model capacity and prediction accuracy, so it can be applied when necessary.

% \begin{figure}[H]
%     \begin{center}
%         \includegraphics[scale=0.45]{img/cascade_simple_hierarchy.pdf}
%     \end{center}
%     \caption{Two possible types of hierarchical CDT architectures.}
%     \label{fig:cdt_hierarchy}
% \end{figure}

% \subsection{Soft Decision Forest}
% Soft decision forest with soft fusion on trees to improve model accuracy but sacrifice the interpretability. Several SDTs can be combined by either simply averaging them or with a weighted combination. There are learnable weights for the approach of weighted combination of SDTs, which will improve the expressivity of model. 

% % Comparison of soft decision forest with averaging or learnable combinations.

% The extension from SDT to SDF is trivial, no matter the trees are averaged or fused with learnable parameters. Suppose that the weight parameter for each tree is $a_k$, where $k=1,2,...,M$ is the index of the tree in a SDF of $M$ trees. Given the importance vectors for each tree $\boldsymbol{I_k}$, we have the overall importance for the SDF as: 
% \begin{equation}
%     \boldsymbol{I} = \sum^M_{k=1} a_k \boldsymbol{I_k}
% \end{equation}

% \subsection{Trade-off between Accuracy and Interpretability}

% Problem of imitating an agent with stochastic policy like PPO: if the PPO converges to a certain policy but still with stochastic policy, it will generate different possible actions based on the same input state, which makes imitation learning not feasible. So normally we choose the imitator to be stochastic as well, and this makes it improper to evaluate the agent with metric like prediction accuracy.

\section{Experiments}
We compare CDT and SDT on two settings for interpreting RL agents: (1) the imitation learning setting, where the RL agent parameterized with a black-box model (\emph{e.g.} neural network) first generates a state-action dataset for imitators to learn from, and the interpretation is derived on the imitators; (2) the full RL setting, where the RL agent is directly trained with the policy represented with interpretable models like CDTs or SDTs, such that the interpretation can be derived by directly spying into those models. The environments are \textit{CartPole-v1}, \textit{LunarLander-v2} and \textit{MountainCar-v0} in OpenAI Gym~\citep{brockman2016openai}. The depth of CDT is represented as "$d_1+d_2$" in the following sections, where $d_1$ is the depth of feature learning tree $\mathcal{F}$ and $d_2$ is the depth of decision making tree $\mathcal{D}$. Each setting is trained for five runs in imitation learning and three runs in RL.

Both the fidelity and stability of mimic models reflect the reliability of them as interpretable models. Fidelity is the accuracy of the mimic model, \emph{w.r.t.} the original model. It is an estimation of similarity between the mimic model and the original one in terms of prediction results. However, fidelity is not sufficient for reliable interpretations.  An unstable family of mimic models will lead to inconsistent explanations of original black-box models. The stability of the mimic model is a deeper excavation into the model itself and comparisons among several runs. Previous research~\citep{bastani2017interpreting} has investigated the fidelity and stability of decision trees as mimic models, where the stability is estimated with the fraction of equivalent nodes in different random decision trees trained under the same settings. In our experiments, the stability analysis is conducted via comparing tree weights of different instances in imitation learning settings.

\subsection{Imitation Learning}
\textbf{Performance.} The datasets for imitation learning are generated with heuristic agents for environments \textit{CartPole-v1} and \textit{LunarLander-v2}, containing 10000 episodes of state-action data for each environments. See Appendix C for other training details. The results are provided in Table~\ref{tab:compare_cdt_sdt1} and \ref{tab:compare_cdt_sdt2}.
% \begin{table}[H]
% \footnotesize
% \centering
% \begin{tabular}{|c|c|c|c|c|}
% \hline
% Tree Type & Depth & Discretized & Accuracy (\%)  & \# of Parameters  \\ \hline
% \multirow{ 6}{*}{SDT} & \multirow{ 2}{*}{2} & \xmark & 94.1\tiny{$\pm0.01$}    & 23\\  \cline{3-5}
% &&\cmark & 49.7\tiny{$\pm0.02$}    & 14\\ \cline{2-5}
% & \multirow{ 2}{*}{3} & \xmark  & 94.5\tiny{$\pm0.1$}    & 51  \\ \cline{3-5}
% &&\cmark & 50.0 \tiny{$\pm0.01$}    & 30\\ \cline{2-5}
% & \multirow{ 2}{*}{4} & \xmark  & 94.3\tiny{$\pm0.3$}  & 107\\  \cline{3-5}
% &&\cmark & 50.1 \tiny{$\pm0.1$}    & 62 \\  \hline

% \multirow{ 6}{*}{CDT (ours)} & \multirow{ 2}{*}{1+2} & \xmark & 95.4\tiny{$\pm1.1$}    & 38\\  \cline{3-5}
% &&\cmark & 94.4\tiny{$\pm0.8$} (84.1\tiny{$\pm2.8$}, 83.8\tiny{$\pm2.6$})   & 35\\ \cline{2-5}
% & \multirow{ 2}{*}{2+1} & \xmark  & 95.6\tiny{$\pm0.1$}    & 54  \\  \cline{3-5}
% &&\cmark & 92.7\tiny{$\pm0.4$} (88.4\tiny{$\pm1.3$}, 89.0\tiny{$\pm0.4$})   & 45\\ \cline{2-5}
% & \multirow{ 2}{*}{2+2} & \xmark  & 96.6\tiny{$\pm0.9$}    & 64\\  \cline{3-5}
% &&\cmark & 91.6\tiny{$\pm1.3$} (82.9\tiny{$\pm3.7$}, 81.9\tiny{$\pm1.8$})   & 55\\  \hline
% \end{tabular}
% \caption{Comparison of CDT and SDT with imitation-learning settings on \textit{CartPole-v1}. In the column of `Accuracy', for discretized CDTs, values in the column are calculated with discretization for the feature learning tree; while values in brackets are discretized CDTs with only discretization for the decision making tree and discretization for both sub-trees.}
% \label{tab:compare_cdt_sdt1}
% \end{table}

\begin{table}[htbp]
% \footnotesize
\scriptsize
\centering
\begin{tabular}{|p{7mm}|p{4mm}|p{10mm}|p{14mm}|p{16mm}|p{6mm}|}
\hline
Tree Type & Depth & Discretized & Accuracy (\%)  & Episode Reward & \# of Params  \\ \hline
\multirow{ 6}{*}{SDT} & \multirow{ 2}{*}{2} &  \xmark  & 94.1\tiny{$\pm0.01$} & 500.0\tiny{$\pm0.0$}  & 23\\  \cline{3-6}
&&\cmark  & 49.7\tiny{$\pm0.02$}  & 39.9\tiny{$\pm7.6$} & 14\\ \cline{2-6}
& \multirow{ 2}{*}{3} & \xmark  & 94.5\tiny{$\pm0.1$} & 500.0\tiny{$\pm0.0$} & 51  \\ \cline{3-6}
&&\cmark  & 50.0 \tiny{$\pm0.01$}  & 42.5\tiny{$\pm7.3$} & 30\\ \cline{2-6}
& \multirow{ 2}{*}{4} & \xmark  & 94.3\tiny{$\pm0.3$} & 500.0\tiny{$\pm0.0$}  & 107\\  \cline{3-6}
&&\cmark  & 50.1 \tiny{$\pm0.1$}   & 40.4\tiny{$\pm7.8$} & 62 \\  \hline

\multirow{ 12}{*}{\textbf{CDT}} & \multirow{ 4}{*}{1+2} & \xmark & \textbf{95.4}\tiny{$\pm1.1$} & \textbf{500.0}\tiny{$\pm0.0$}  & 38\\  \cline{3-6}
&&$\mathcal{F}$ only & \textbf{94.4}\tiny{$\pm0.8$} &  \textbf{500.0}\tiny{$\pm0.0$} & 35\\ \cline{3-6}
&&$\mathcal{D}$ only & 84.1\tiny{$\pm2.8$} &  500.0\tiny{$\pm0.0$} & 35\\ \cline{3-6}
&&$\mathcal{F} + \mathcal{D}$ & \textbf{83.8}\tiny{$\pm2.6$} & 497.8\tiny{$\pm8.4$} & \textbf{32}\\ \cline{2-6}
& \multirow{ 4}{*}{2+1} & \xmark  & \textbf{95.6}\tiny{$\pm0.1$}  & \textbf{500.0}\tiny{$\pm0.0$} & 54  \\  \cline{3-6}
&& $\mathcal{F}$ only & \textbf{92.7}\tiny{$\pm0.4$} & \textbf{500.0}\tiny{$\pm0.0$} & 45\\ \cline{3-6}
&& $\mathcal{D}$ only & 88.4\tiny{$\pm1.3$} & 500.0\tiny{$\pm0.0$} & 53\\ \cline{3-6}
&& $\mathcal{F} + \mathcal{D}$ & \textbf{89.0}\tiny{$\pm0.4$} & \textbf{500.0}\tiny{$\pm0.0$} & \textbf{44}\\ \cline{2-6}
& \multirow{ 4}{*}{2+2} & \xmark  & \textbf{96.6}\tiny{$\pm0.9$} &  \textbf{500.0}\tiny{$\pm0.0$} & 64\\  \cline{3-6}
&&$\mathcal{F}$ only & 91.6\tiny{$\pm1.3$}  & 500.0\tiny{$\pm0.0$} & 55\\  \cline{3-6}
&&$\mathcal{D}$ only & 82.9\tiny{$\pm3.7$}  & 494.8\tiny{$\pm19.8$} & 61\\  \cline{3-6}
&&$\mathcal{F} + \mathcal{D}$ & 81.9\tiny{$\pm1.8$} & 488.8\tiny{$\pm31.4$} & 52\\  \hline

\end{tabular}
\caption{Comparison of CDT and SDT with imitation-learning settings on \textit{CartPole-v1}.}
\label{tab:compare_cdt_sdt1}
\end{table}

CDTs perform consistently better than SDTs before and after discretization process in terms of prediction accuracy, with different depths of the tree. Additionally, for providing a similarly accurate model, CDT method always has a much smaller number of parameters compared with SDT, which improves its interpretability as shown in later sections. However, although better than SDTs, CDTs also suffer from degradation in performance after discretization, which could lead to unstable and unexpected models. We claim that this is a general drawback for tree-based methods with soft decision boundaries in XRL with imitation-learning settings, which is further studied in the following.

\begin{table}[htbp]
% \footnotesize
\scriptsize
\centering
\begin{tabular}{|p{7mm}|p{4mm}|p{10mm}|p{14mm}|p{16mm}|p{6mm}|}
\hline
Tree Type & Depth & Discretized & Accuracy (\%) & Episode Reward & \# of Params  \\ \hline
\multirow{ 8}{*}{SDT} & \multirow{ 2}{*}{4} & \xmark & 85.4\tiny{$\pm0.4$} & 58.2\tiny{$\pm246.1$} & 199\\  \cline{3-6}
&&\cmark & 54.8\tiny{$\pm10.1$}   & -237.1\tiny{$\pm121.9$} & 94\\ \cline{2-6}
& \multirow{ 2}{*}{5} & \xmark & 87.6\tiny{$\pm0.5$}    & 191.3\tiny{$\pm143.8$} & 407 \\ \cline{3-6}
&&\cmark & 51.6\tiny{$\pm4.5$}   & -93.7\tiny{$\pm102.9$} & 190\\ \cline{2-6}
& \multirow{ 2}{*}{6} & \xmark  & 88.7\tiny{$\pm1.3$}  & 193.4\tiny{$\pm161.4$} & 823\\  \cline{3-6}
&&\cmark & 60.2\tiny{$\pm3.9$}   & -172.4\tiny{$\pm122.0$} & 382\\ \cline{2-6}
& \multirow{ 2}{*}{7} & \xmark   & 88.9\tiny{$\pm0.5$}   & 194.2\tiny{$\pm138.8$} & 1655\\  \cline{3-6}
&&\cmark & 62.7\tiny{$\pm2.8$}   & -233.4\tiny{$\pm62.4$} & 766\\ \hline

\multirow{ 16}{*}{\textbf{CDT}} & \multirow{ 4}{*}{2+2} & \xmark  & \textbf{88.2}\tiny{$\pm1.6$}   & \textbf{107.4}\tiny{$\pm190.7$}  & \textbf{116}\\  \cline{3-6}
&&$\mathcal{F}$ only & \textbf{78.0}\tiny{$\pm2.4$}  & -126.9\tiny{$\pm237.0$}  & \textbf{95}\\ \cline{3-6}
&&$\mathcal{D}$ only & 68.3\tiny{$\pm10.3$}  & -301.6\tiny{$\pm136.8$}  & 113\\ \cline{3-6}
&&$\mathcal{F}+\mathcal{D}$ & \textbf{64.4}\tiny{$\pm12.1$}  & -229.7\tiny{$\pm256.0$}  & \textbf{92}\\ \cline{2-6}
& \multirow{ 4}{*}{2+3} & \xmark  & \textbf{88.3}\tiny{$\pm1.7$}  & \textbf{168.5}\tiny{$\pm169.0$}  & \textbf{144} \\  \cline{3-6}
&&$\mathcal{F}$ only & 70.2\tiny{$\pm2.3$}   & -9.7\tiny{$\pm159.2$}  & 123\\ \cline{3-6}
&&$\mathcal{D}$ only & 40.7\tiny{$\pm11.9$}  & -106.3\tiny{$\pm187.7$}  & 137\\ \cline{3-6}
&&$\mathcal{F}+\mathcal{D}$ & 35.9\tiny{$\pm1.5$}  & -130.2\tiny{$\pm135.9$}  & 116\\ \cline{2-6}
& \multirow{ 4}{*}{3+2} & \xmark & \textbf{90.4}\tiny{$\pm1.7$}   & \textbf{199.5}\tiny{$\pm123.7$}  & 216\\  \cline{3-6}
&&$\mathcal{F}$ only & 72.2\tiny{$\pm8.3$}   & -14.2\tiny{$\pm175.6$}  & 167\\ \cline{3-6}
&&$\mathcal{D}$ only  & \textbf{78.1}\tiny{$\pm2.5$}  & \textbf{150.8}\tiny{$\pm148.1$}  & 209\\ \cline{3-6}
&&$\mathcal{F}+\mathcal{D}$ & \textbf{64.6}\tiny{$\pm4.7$}  & \textbf{7.1}\tiny{$\pm173.6$}  & \textbf{160}\\ \cline{2-6}
& \multirow{ 4}{*}{3+3} & \xmark  & \textbf{90.4}\tiny{$\pm1.2$}   & \textbf{173.0}\tiny{$\pm124.5$}  & 244\\  \cline{3-6}
&&$\mathcal{F}$ only & 72.0\tiny{$\pm1.2$}   & -55.3\tiny{$\pm178.6$}  & 195\\ \cline{3-6}
&&$\mathcal{D}$ only & 58.7\tiny{$\pm8.6$}  & -91.5\tiny{$\pm97.0$}  & 237 \\ \cline{3-6}
&&$\mathcal{F}+\mathcal{D}$ & 46.8\tiny{$\pm5.6$}  & -210.5\tiny{$\pm121.9$}  & 188 \\ \hline

\end{tabular}
\caption{Comparison of CDT and SDT with imitation-learning settings on \textit{LunarLander-v2}.}
\label{tab:compare_cdt_sdt2}
\end{table}

\textbf{Stability.}
\label{sec:stablility}
To investigate the stability of imitation learners for interpreting the original agents, we measure the normalized weight vectors from different imitation-learning trees. For SDTs, the weight vectors are the linear weights on inner nodes, while for CDTs $\{\tilde{\boldsymbol{w}}, \tilde{\boldsymbol{w}}^\prime \}$ are considered. Through the experiments, we would like to show how unstable the imitators $\{\mathbf{L}\}$ are. We have a tree agent $\mathbf{X}\in\{\mathbf{L'}, \mathbf{H}, \mathbf{R}\}$, where $\mathbf{L'}$ is another imitator tree agent trained under the same setting, $\mathbf{R}$ is a random tree agent, and $\mathbf{H}$ is a heuristic tree agent (used for generating the training dataset). The distances of tree weights between two agents $\mathbf{L, X}$ are calculated with the following formula:
\begin{align*}
    D(\mathbf{L}, \mathbf{X}) &= \frac{1}{2N}\sum^N_{n=1} \min_{m=1,2,...,M} || \boldsymbol{l}_m-\boldsymbol{x}_n ||_1 \\
    &+ \frac{1}{2M}\sum^M_{m=1} \min_{n=1,2,...,N} || \boldsymbol{x}_m-\boldsymbol{l}_n ||_1  \numberthis
\end{align*}
where $M$ is the number of the imitation learners, and $N$ is the number of another set of imitation learners ($\mathbf{L'}$), or the number of heuristic agents ($\mathbf{H}$), or the number of random tree agents ($\mathbf{R}$), depending on the specific choice of $\mathbf{X}$. $\overline{D(\mathbf{L}, \mathbf{X})}$ are averaged over all possible $\mathbf{L}$s and $\mathbf{X}$s with the same setting.
Since we have the heuristic agent for \textit{LunarLander-v2} environment and we transform the heuristic agent into a multivariate DT agent, we get the decision boundaries of the tree on all its nodes. So we also compare the differences of decision boundaries in heuristic tree agent $\mathbf{H}$ and those of the learned tree agent $\mathbf{L}$. But we do not have the official heuristic agent for \textit{CartPole-v1} in the form of a decision tree. For the decision making trees in CDTs, we transform the weights back into the input feature space to make a fair comparison with SDT and the heuristic tree agent.  The results are displayed in Table~\ref{tab:stability}, all trees use intermediate features of dimension 2 for both environments. In terms of stability, CDTs generally perform similarly as SDTs and even better on \textit{CartPole-v1} environment.
% Specifically, for normalized decision boundary vectors $\{\boldsymbol{h}_m| m=1,2,...,M\}$ in heuristic agent and $\{\boldsymbol{l}_n| n=1,2,...,N\}$ in SDT, the difference is evaluated with the following metric:
% \begin{equation}
%     D(\mathbf{H}, \mathbf{L}) = \frac{1}{N}\sum^N_{n=1} \min_{m=1,2,...,M} || \boldsymbol{h}_m-\boldsymbol{l}_n ||_1
% \end{equation}

\begin{table}
% \footnotesize
\scriptsize
\centering
\begin{tabular}{ |p{5mm}|p{14.5mm}|p{8mm}|p{9mm}|p{12mm}|p{9mm}| } 
% \begin{tabular}{ m{1cm} m{1cm}  m{1cm} m{1cm} m{1cm} m{1cm} } 
 \hline
 Tree Type & Env & Depth &   $\overline{D(\mathbf{L}, \mathbf{L'})}$ &  $\overline{D(\mathbf{L}, \mathbf{R})}$ &  $\overline{D(\mathbf{L}, \mathbf{H})}$ \\[0.8ex] \hline 
 \multirow{ 2}{*}{SDT} & CartPole-v1 & 3  & 0.21 & 0.90$\pm 0.10$ & - \\ \cline{2-6}
 & \multirow{1}{*}{LunarLander-v2} & 4  & 0.50 & 0.92$\pm 0.05$ & 0.84\\ \hline
 
 \multirow{ 4}{*}{\textbf{CDT}} & \multirow{ 2}{*}{CartPole-v1} & 1+2  & 0.07 & 1.05$\pm0.15$ & - \\ \cline{3-6}
 & & 2+2  & 0.19 & 1.03$\pm0.10$ & - \\ \cline{2-6}
  & \multirow{ 2}{*}{LunarLander-v2}  & 2+2  & 0.63 & 1.01$\pm0.10$ & 0.98\\ \cline{3-6} 
 & & 3+3 & 0.53 & 0.83$\pm0.06$ & 0.86 \\ 
 \hline
\end{tabular}
\caption{Tree Stability Analysis. $\overline{D(\mathbf{L}, \mathbf{L'})}$, $\overline{D(\mathbf{L}, \mathbf{R})}$ and $\overline{D(\mathbf{L}, \mathbf{H})}$ are average values of distance between an imitator $\mathbf{L}$ and another imitator $\mathbf{L}^\prime$, or a random agent $\mathbf{R}$, or a heuristic agent $\mathbf{H}$ with metric $D$. CDTs are generally more stable, but still with large variances over different imitators.}
\label{tab:stability}
\end{table}

% \begin{center}
% \begin{tabular}{ |c|c|c|c|c| } 
%  \hline
%  cell1 & $\overline{D(\mathbf{L}, \mathbf{L'})}$ & Normalized $\overline{D(\mathbf{L}, \mathbf{L'})}$ & Normalized $\overline{D(\mathbf{L}, \mathbf{H})}$ & Normalized $\overline{D(\mathbf{L}, \mathbf{R})}$ \\ \hline
%  CartPole-v1 & 23.9 & 0.21 & - & 0.90 \\ \hline
%  LunarLander-v2 & 25.7 & 0.50 & 0.84 & 0.92\\ 
%  \hline
% \end{tabular}
% \end{center}

We further evaluate the feature importance with at least two different methods on SDTs to demonstrate the instability of imitation learning settings for XRL, see Appendix D.1. We also display all trees (CDTs and SDTs) for both environments in Appendix D.3. Significant differences can be found in different runs for the same tree structure with the same training setting, which testifies the unstable and unrepeatable nature by interpreting imitators instead of the original agents.

% We display the agents trained with CDTs and SDTs on \textit{LunarLander-v2} as an example. Fig.~\ref{fig:sdt_vis} and Fig.~\ref{fig:cdt_vis} show the imitators with SDT and CDT respectively. Each figure contains trees trained in four runs under the same settings. Each sub-figure contains one learned tree with an inference path (in bold) for the same input. The colors of the squares on tree nodes show the values of weight vectors for each node. For feature learning trees in CDTs, the leaf nodes are colored with the feature coefficients. The output leaf nodes are colored with the output categorical distributions. 

% \begin{figure}[H]
%     \begin{center}
%         \includegraphics[scale=0.2]{img/sdt_vis.pdf}
%     \end{center}
%     \caption{Comparison of four runs with the same setting for SDT (before discretization) imitation learning.}
%     \label{fig:sdt_vis}
% \end{figure}

% \begin{figure}[H]
%     \begin{center}
%         \includegraphics[scale=0.3]{img/cdt_fl_vis.pdf} \\
%         \includegraphics[scale=0.3]{img/cdt_dm_vis.pdf}
%     \end{center}
%     \caption{Comparison of four runs with the same setting for CDT (before discretization) imitation learning: feature learning trees (top) and decision making trees (bottom).}
%     \label{fig:cdt_vis}
% \end{figure}

% Other plots including the tree structures after discretization and for \textit{CartPole-v1} environment are provided in Appendix~\ref{app:tree_structures}.

% Conclusion
\textbf{Conclusion.} We claim that the current imitation-learning setting with tree-based models is not suitable for interpreting the original RL agent, with the following evidence derived from our experiments: (i) The discretization process usually degrades the performance (prediction accuracy) of the agent significantly, especially for SDTs. Although CDTs alleviate the problem to a certain extent, the performance degradation is still not negligible, therefore the imitators are not expected to be alternatives for interpreting the original agents; (ii) With the stability analysis in our experiments, we find that different imitators will display different tree structures even if they follow the same training setting on the same dataset, which leads to significantly different decision paths and local feature importance assignments.

\subsection{Reinforcement Learning}

\begin{figure}[htbp]
    \centering
        \centering\includegraphics[scale=0.3]{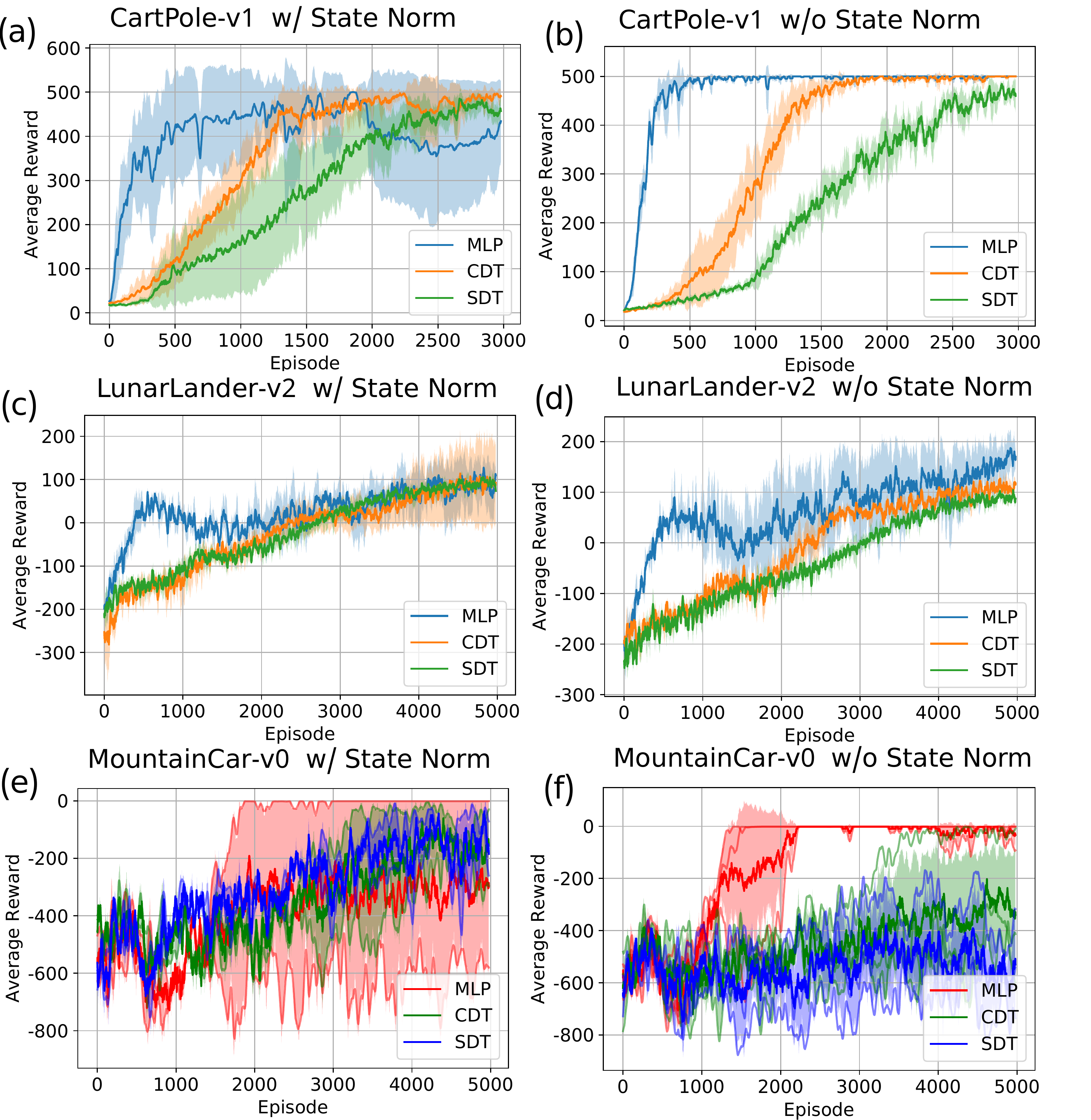}
    \caption{Comparison of SDTs and CDTs on three environments in terms of average rewards in RL setting: (a)(c)(e) use normalized input states while (b)(d)(f) use unnormalized ones.}
    \label{fig:rl_compare}
\end{figure}

\textbf{Performance.} We evaluate the learning performances of different DTs and NNs as policy function approximators in RL, as shown in Fig.~\ref{fig:rl_compare}. Every setting is trained for three runs. We use Proximal Policy Optimization~\citep{schulman2017proximal} algorithm in our experiments. The multilayer perceptron (MLP) model is a two-layer NN with 128 hidden units. The SDT has a depth of 3 for \textit{CartPole-v1} and 4 for \textit{LunarLander-v2}. The CDT has depths of 2 and 2 for feature learning tree and decision making tree respectively on \textit{CartPole-v1}, while with depths of 3 and 3 for \textit{LunarLander-v2}. Therefore for each environment, the SDTs and CDTs have a similar number of model parameters, while MLP model has at least 6 times more parameters. Detailed training settings are provided in Appendix E. From Fig.~\ref{fig:rl_compare}, we can see that CDTs can at least outperform SDTs as policy function approximators for RL in terms of both sampling efficiency and final performance, although may not learn as fast as general MLPs with a significantly larger number of parameters.  For \textit{MountainCar-v0} environment, the MLP model has two layers with 32 hidden units. The depth of SDT is 3. CDT has depths 2 and 2 for the feature learning tree and decision making tree respectively, with the dimension of the intermediate feature as 1. The learning performances are less stable due to the sparse reward signals and large variances in exploration. However, with CDT for policy function approximation, there are still near-optimal agents after training with or without state normalization.

\begin{figure}[H]
    \centering
        \centering\includegraphics[scale=0.29]{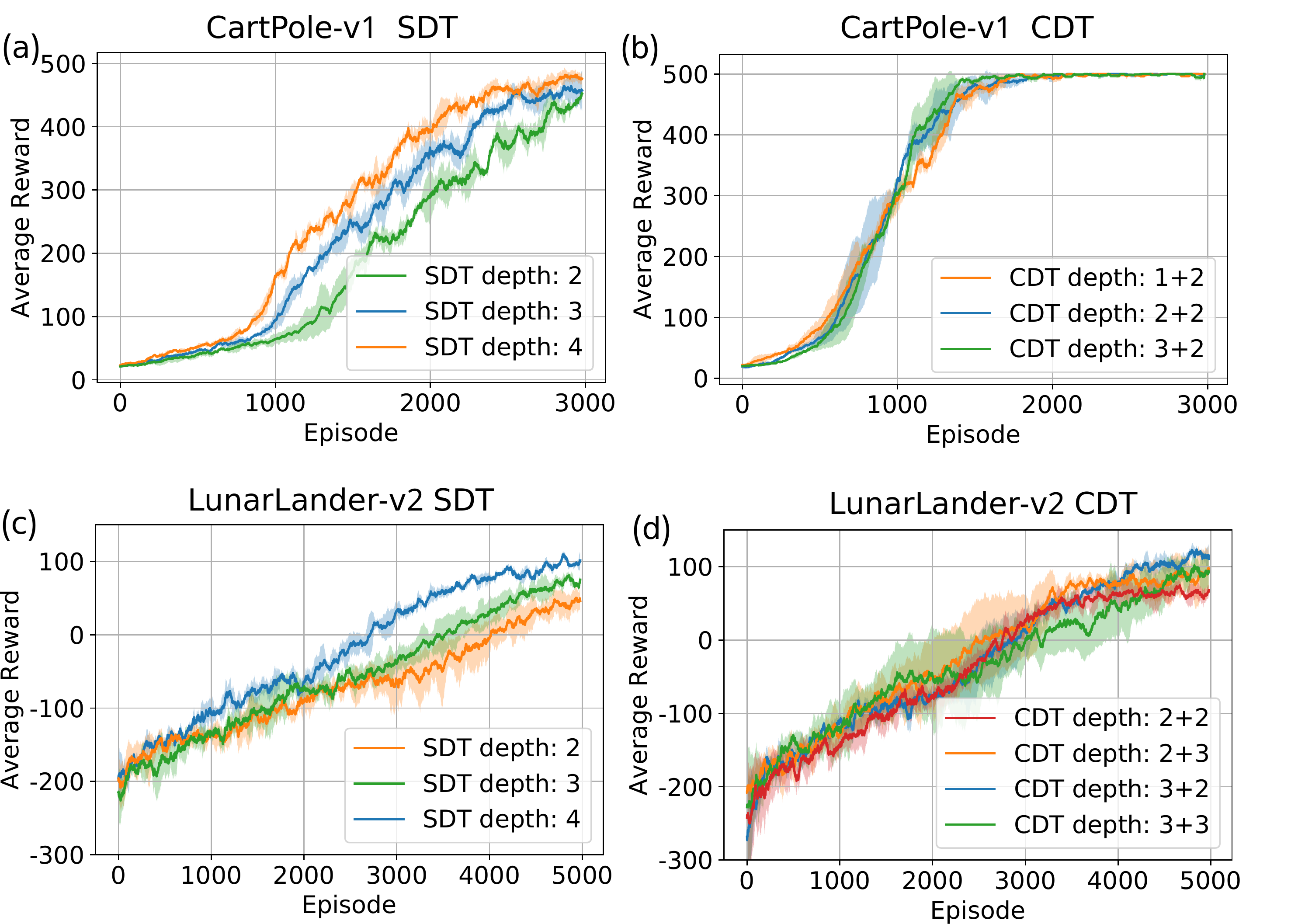}
    \caption{Comparison of SDTs and CDTs with different depths (state normalized). (a) and (b) are trained on \textit{CartPole-v1}, while (c) and (d) are on \textit{LunarLander-v2}.}
    \label{fig:rl_compare_depth_norm}
\end{figure}

\textbf{Tree Depth.} The depths of DTs are also investigated for both SDT and CDT, because deeper trees tend to have more model parameters and therefore lay more stress on the accuracy rather than interpretability. Fig.~\ref{fig:rl_compare_depth_norm} shows the learning curves of SDTs and CDTs in RL with different tree depths for the two environments, using normalized states as input, while the comparison with unnormalized states is in Appendix F with similar results. From the comparisons, we can see that generally deeper trees can learn faster with even better final performances for both CDTs and SDTs, but CDTs are less sensitive to tree depth than SDTs.

\begin{figure}[htbp]
    \centering
        \centering
        \includegraphics[scale=0.16]{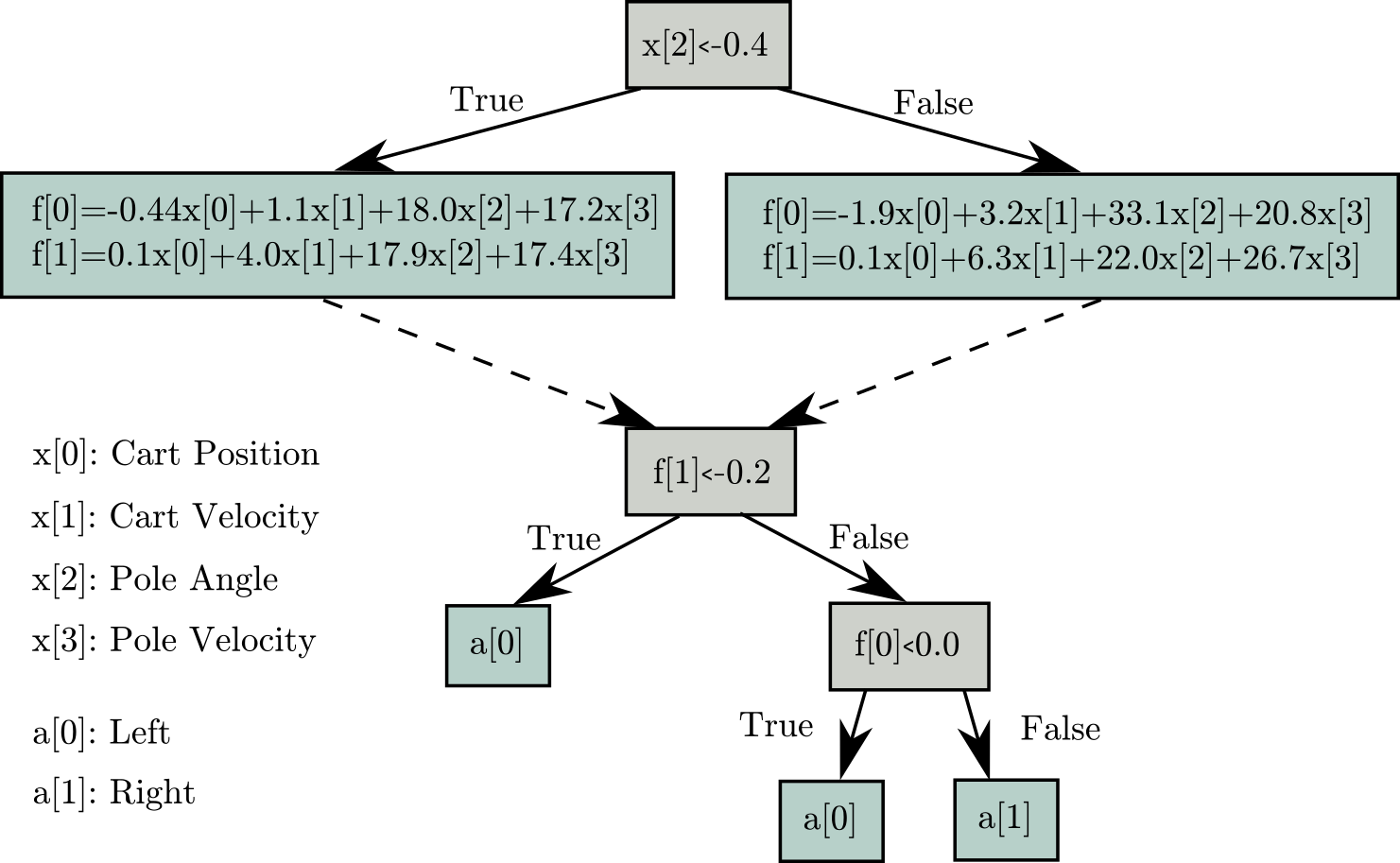}
        \hspace{1pt}
        \includegraphics[scale=0.4]{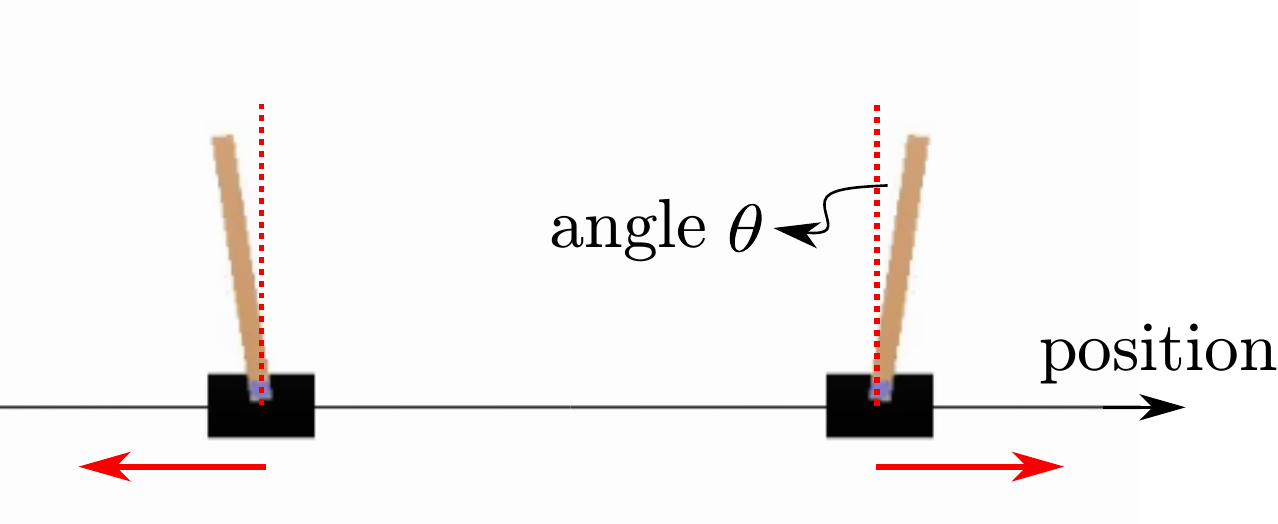}
    \caption{Top: learned CDT (after discretization). Bottom: game scene of \textit{CartPole-v1}. }
    \label{fig:cartpole_plot}
\end{figure}

\textbf{Interpretability.} We display the learned CDTs in RL settings for three environments, compared against some heuristic solutions or SDTs.
A heuristic solution\footnote{Provided by Zhiqing Xiao on OpenAI Gym Leaderboard: https://github.com/openai/gym/wiki/Leaderboard} for \textit{CartPole-v1} is: if $3\theta+\dot{\theta}>0$, push right; otherwise, push left. As shown in Fig.~\ref{fig:cartpole_plot}, in our learned CDT of depth 1+2, the weights of two-dimensional intermediate features ($f[0]$ and $f[1]$) are much larger on the last two dimensions of observation than the first two, therefore we can approximately ignore the first two dimensions due to their low importance in decision making process. So we get similar intermediate features for two cases in two dimensions, which are approximately $w_1x[2]+w_2x[3]\rightarrow w\theta + \dot{\theta}$ after normalization $(w>0)$. Based on the decision making tree in learned CDT, it gives a close solution as the heuristic one, yielding if $w\theta+\dot{\theta}<0$ push left otherwise push right. The original CDT before discretization and a SDT for comparison are provided in Appendix G.

\begin{figure}[htbp]
    \centering
        \centering
        \includegraphics[scale=0.15]{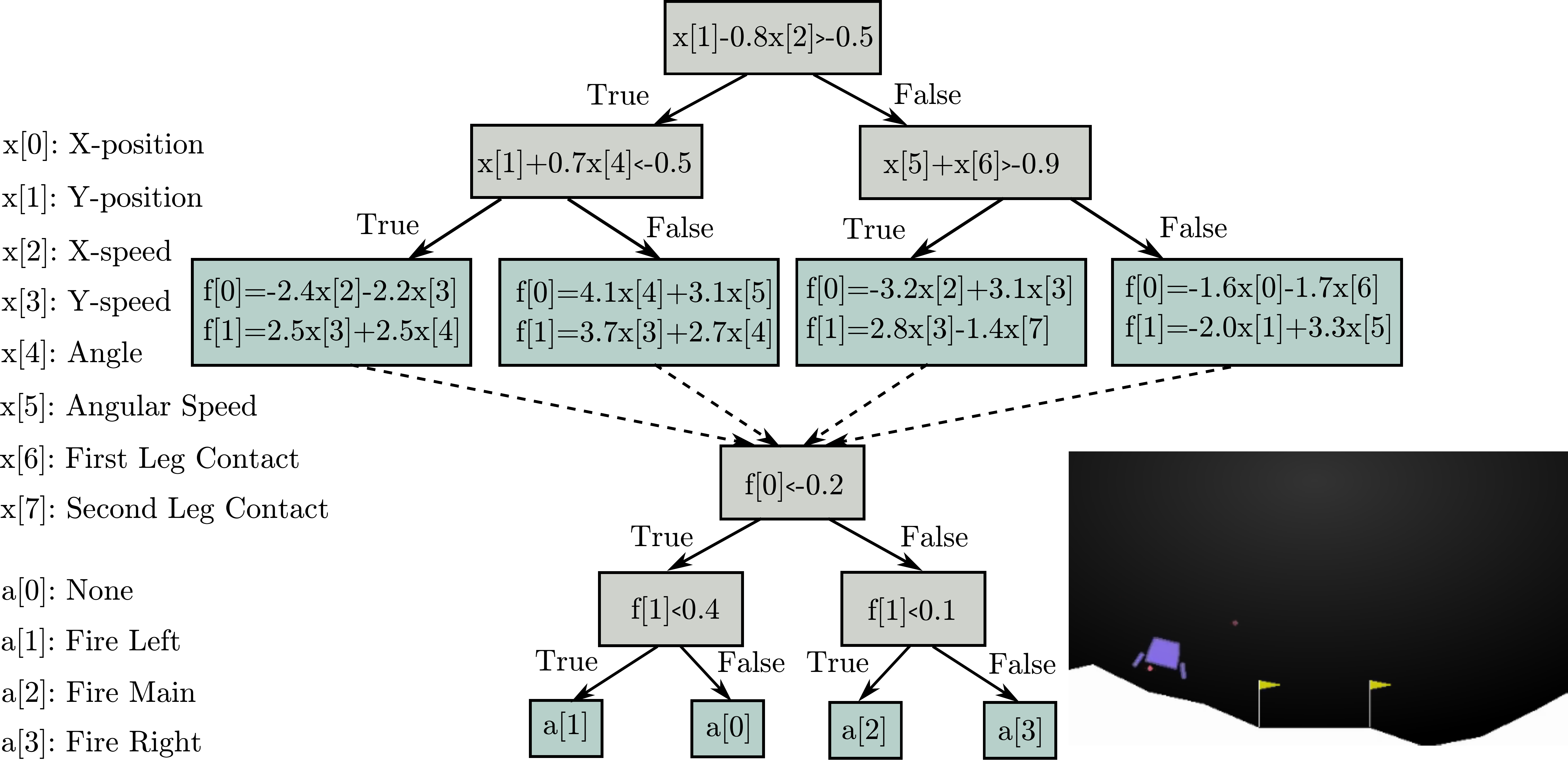}
        % \hspace{50pt}
        % \includegraphics[scale=0.2]{img/lunarlander.png}
    \caption{Learned CDT (after discretization) for \textit{LunarLander-v2} with game scene at the right bottom corner.}
    \label{fig:mountaincar_plot}
\end{figure}

For \textit{MountainCar-v0}, due to the complexity in the landscape as shown in Fig.~\ref{fig:mountaincar_plot}, interpreting the learned model is even harder. However, through CDT, we can see that the agent learns intermediate features as combinations of car position and velocity, potentially being an estimated future position or previous position, and makes action decisions based on that. The original CDT before discretization has depth 2+2 with one-dimensional intermediate features, and its structure is shown in Appendix G.
For \textit{LunarLander-v2}, as in Fig.~\ref{fig:lunarlander_plot}, the learned CDT agent captures some important feature combinations like the angle with angular speed and X-Y coordinate relationships for decision making.

\begin{figure}[htbp]
    \centering
        \centering
        \includegraphics[scale=0.18]{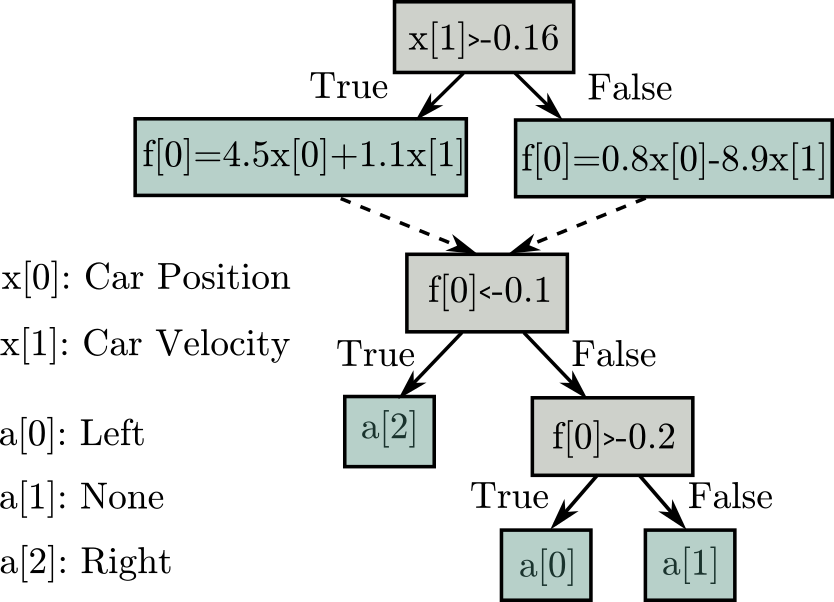}
        % \hspace{50pt}
        \includegraphics[scale=0.2]{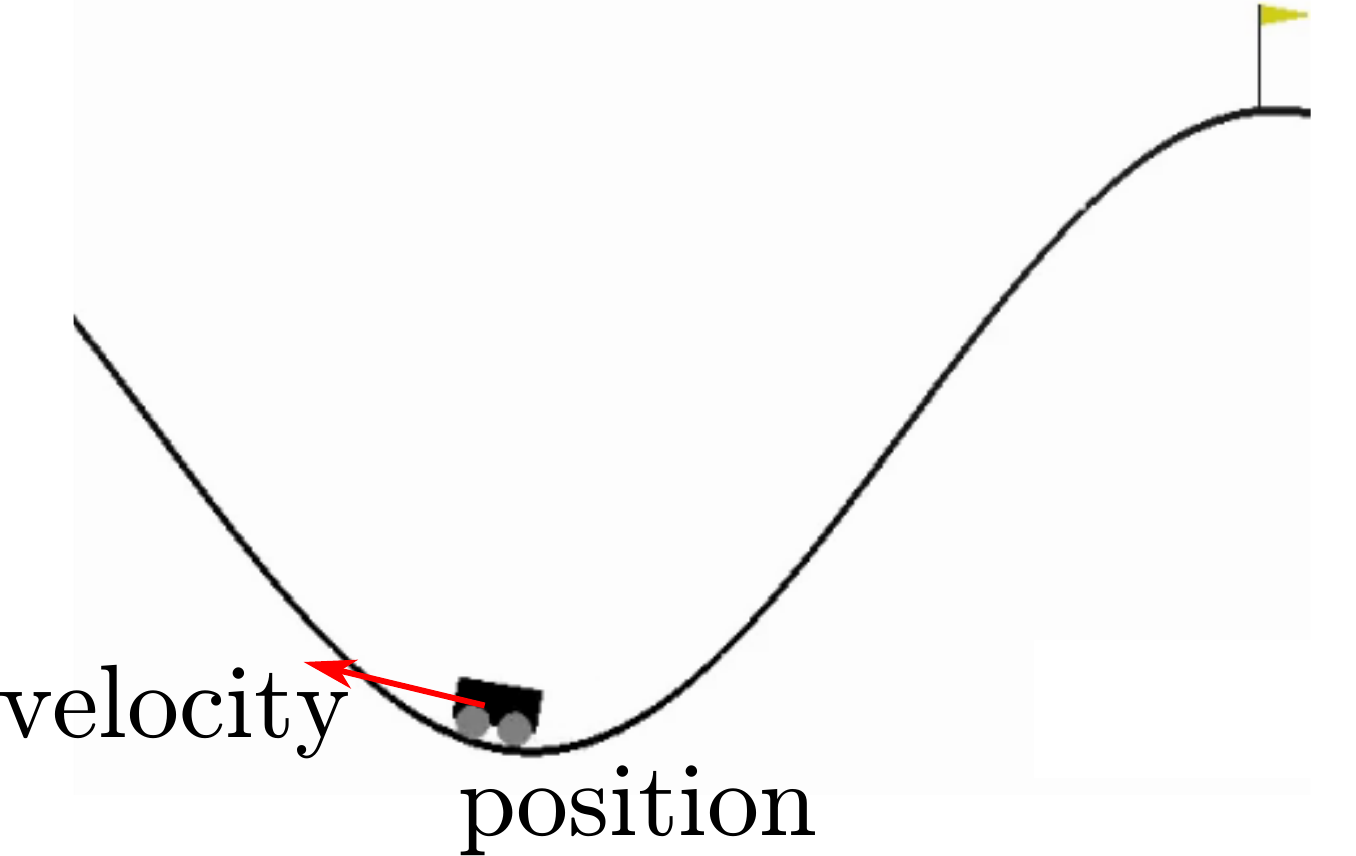}
    \caption{Left: learned CDT (after discretization). Right: game scene of \textit{MountainCar-v0}.}
    \label{fig:lunarlander_plot}
\end{figure}

\section{Conclusion}
% We have proposed in this paper a new architecture of differentiable DT, which is called the CDT. A simple CDT cascades a feature learning DT and a decision making DT as a whole. From our experiments, we demonstrate that compared with traditional differentiable DTs like SDT, CDT has even better function approximation ability in both IL and RL settings with significantly reduced number of model parameters, while better preserves the tree prediction accuracy after discretization; surprisingly, we also qualitatively and quantitively testify that the IL setting may not be proper for achieving interpretable RL agents due to the significant differences and instability among different imitators in their tree structures, although they may have similar performances. Finally, we display the interpretability of learned CDTs with RL settings, especially for the intermediate features. CDT constructs a type of methods that can be further extended to hierarchical architectures with more interpretable modules, giving the credit to its richer expressivity by allowing for representation learning.

In this work, we have proposed a new architecture of differentiable DT, the Cascading Decision Tree (CDT). A simple CDT cascades a feature learning DT and a decision making DT into a single model. From our experiments, we show that compared with traditional differentiable DTs (\emph{i.e.}, DDTs or SDTs) CDTs have better function approximation in both imitation learning and full RL settings with a significantly reduced number of model parameters while better preserving the tree prediction accuracy after discretization. We also qualitatively and quantitively corroborate that the SDT-based methods with imitation learning setting may not be proper for achieving interpretable RL agents due to instability among different imitators in their tree structures, even when having similar performances. Finally, we contrast the interpretability of learned DTs in RL settings, especially for the intermediate features. Our analysis supports that CDTs lend themselves to be further extended to hierarchical architectures with more interpretable modules, due to its richer expressivity allowed via representation learning. More work needs to be done to fully realize the potential of our method, which involves the investigation of hierarchical CDT settings and well-regularized intermediate features for further interpretability. Additionally, since the present experiments are demonstrated with linear transformations in the feature space, non-linear transformations are expected to be leveraged for tasks with higher complexity or continuous action space while preserving interpretability.

% \section{Conclusion}
% Future works:

% from discrete to continuous action space

% soft decision forests

% high-dimensional observation

% more about relationship between learned CDT and HDT

\bibliography{example_paper}
\bibliographystyle{icml2021}

\onecolumn
\begin{center}
    \Large
    \textbf{Supplementary Material}
\end{center}
\begin{appendices}

\section{Detailed Simple CDT Architecture}
\label{subsec:appendix_cdt}
\begin{figure}[H]
    \begin{center}
        \includegraphics[scale=0.45]{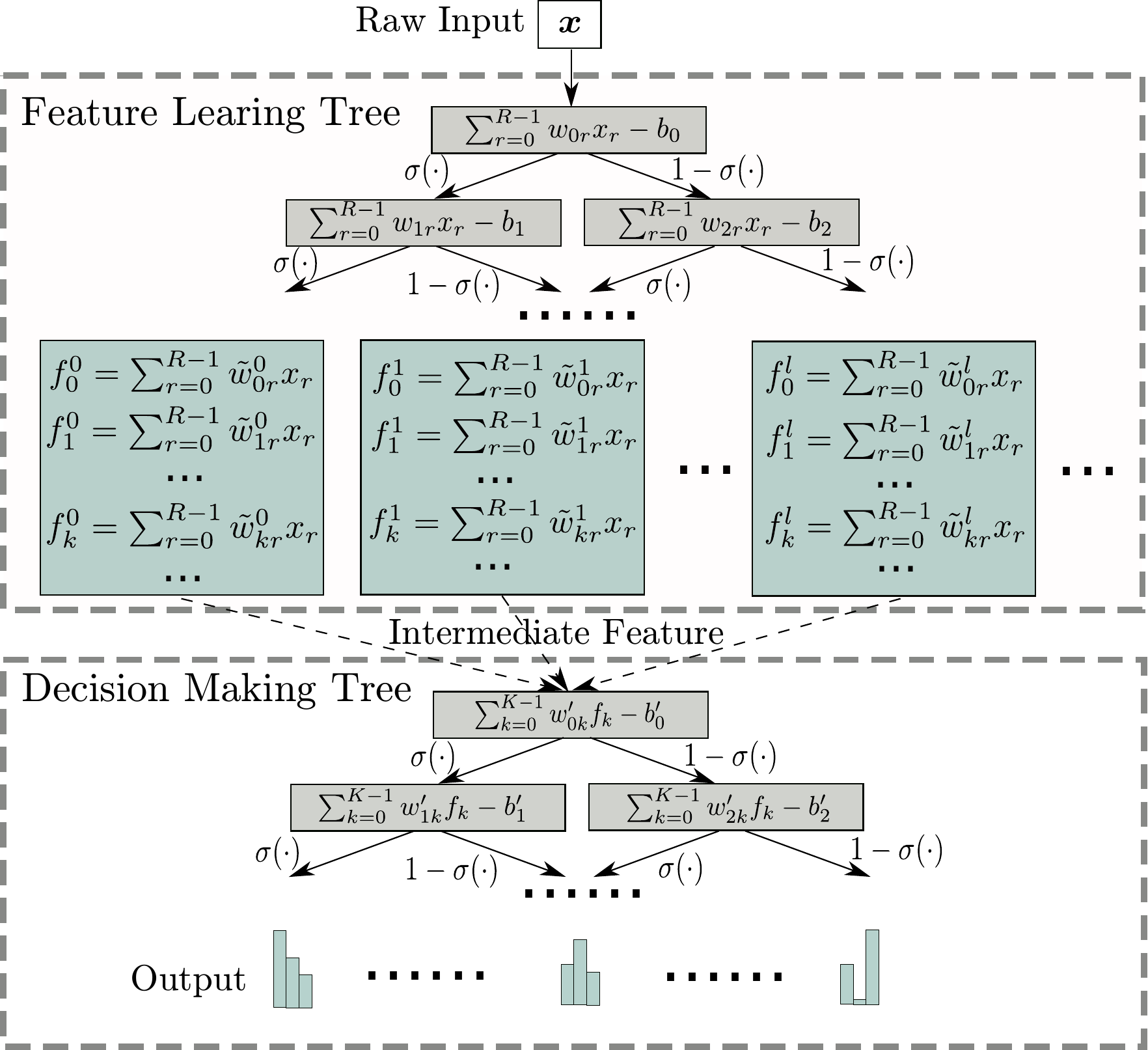}
    \end{center}
    \caption{A detailed architecture of simple CDT: the feature learning tree is parameterized by $w$ and $b$, and its leaves are parameterized by $\tilde{w}$
    % , which represents $L$ possible intermediate feature values with a dimension of $K$
    ; the inner nodes of decision making tree are parameterized by $w^\prime$ and $b^\prime$, while the leaves are parameterized by $\tilde{w}^\prime$.}
    \label{fig:cascade}
\end{figure}

\section{Quantitative Analysis of Model Parameters}
\label{app:model_parameters}
Considering the case where we have a raw feature dimension of inputs as $R$, we choose the intermediate feature dimension to be $K<R$. A CDT with two cascading trees of depth $d_1$ and $d_2$ and a SDT with depth $d$ are compared. Supposing the output dimension is $O$, we can derive the number of parameters in CDT as:
\begin{equation}
    N(CDT)=[(R+1)(2^{d_1}-1)+K\cdot R\cdot2^{d_1}]+[(K+1)(2^{d_2}-1)+O\cdot2^{d_2}]
    \label{equ:cdt_parameters}
\end{equation}
while the number of parameters in SDT is:
\begin{equation}
    N(SDT)=(R+1)(2^d-1)+O\cdot2^d
    \label{equ:sdt_parameters}
\end{equation}

Considering an example for Eq.~\ref{equ:cdt_parameters} and Eq.~\ref{equ:sdt_parameters} with SDT being depth of 5 while CDT has $d_1=2, d_2=3$, raw feature dimension $R=8$, intermediate feature dimension $K=4$, and output dimension $O=4$, we can get $N(CDT)=222$ and $N(SDT)=343$. It indicates a reduction of around $35\%$ parameters in this case, which will significantly increase interpretability.

\begin{figure}[htbp]
    \begin{center}
        \includegraphics[scale=0.6]{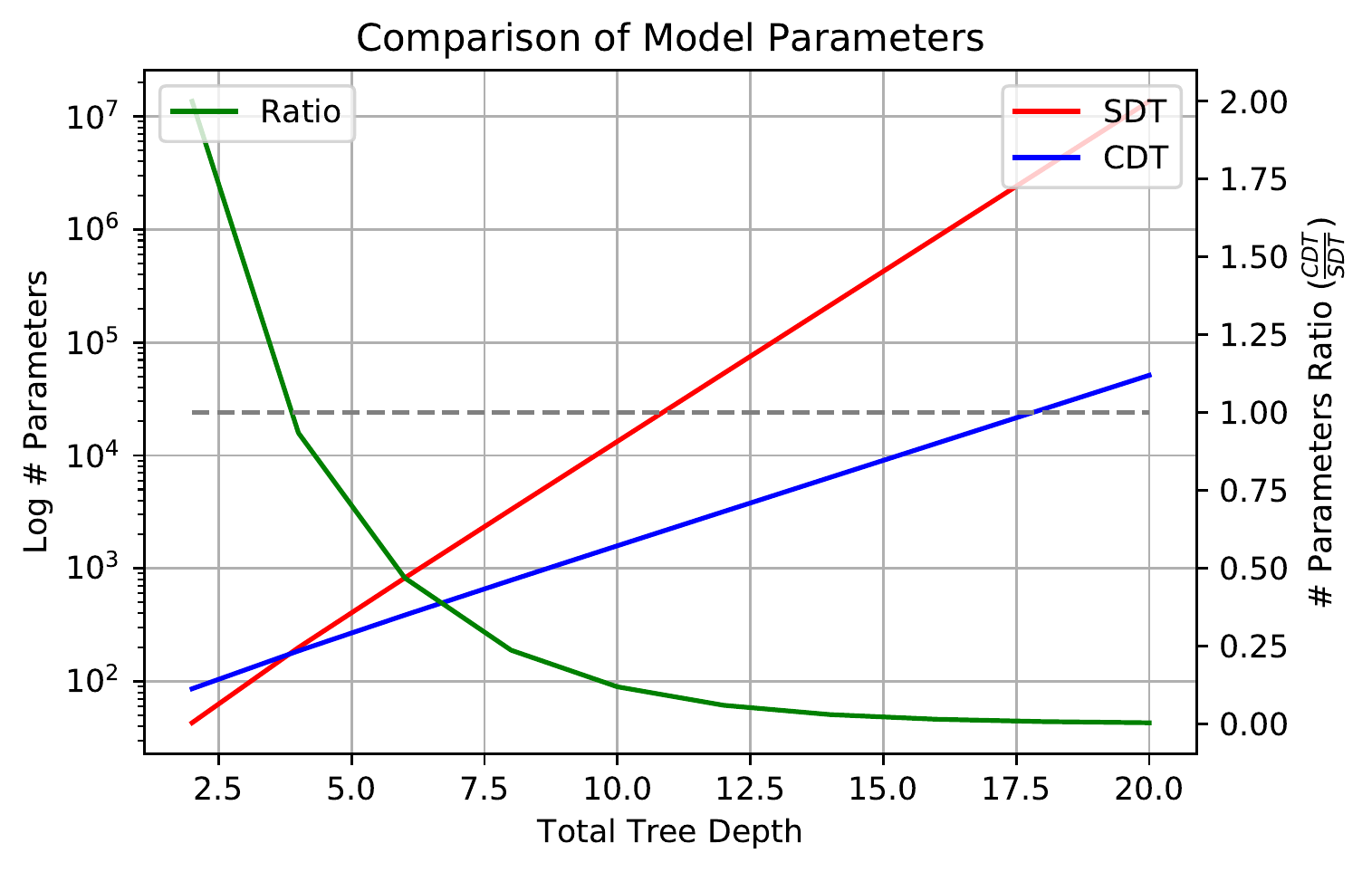}
    \end{center}
    \caption{Comparison of numbers of model parameters in CDTs and SDTs. The left vertical axis is the number of model parameters in $\log$-scale. The right vertical axis is the ratio of model parameter numbers. CDT has a decreasing ratio of model parameters against SDT as the total depth of model increases.  }
    \label{fig:params_comparison}
\end{figure}

In another example, when $R=8, K=4, O=4$, the numbers of parameters in CDT or SDT models are compared in Fig.~\ref{fig:params_comparison}, assuming $d=d_1+d_2$ for a total depth of range 2 to 20. The Ratio of numbers of model parameters is derived with: $\frac{N(CDT)}{N(SDT)}$.

% \section{Confidence Weighted Feature Importance}
% \label{app:weighted_importance}
% \begin{figure}[H]
%     \begin{center}
%         \includegraphics[scale=0.3]{img/boundary2.pdf}
%     \end{center}
%     \caption{Figure shows multiple soft decision boundaries partition the space. The boundaries closer to the instance are less important in determining the feature importance since they are less distinctive for the instance.}
%     \label{fig:boundary2}
% \end{figure}
% Fig.~\ref{fig:boundary2} helps to demonstrate the reason for using the decision confidence (\emph{i.e.}, probability) as a weight for assigning feature importance, which indicates that the probability of belonging to one category is positively correlated with the distance from the instance to the decision boundary. Therefore when there are multiple boundaries for partitioning the space (\emph{e.g.}, two in the figure), we assign the boundaries with shorter distance to the data point with smaller confidence in determining feature importance, since based on the closer boundaries the data point is much easier to be perturbed into the contrary category and less confident to remain in the original.

\section{Hyperparameters in Imitation Learning}
\label{app:il_params}
\begin{table}[H]
\centering
% \begin{tabular}{ |c|c|c|c|c| } 
\begin{tabular}{ m{2cm} m{3cm} m{2cm} m{2cm} } 
 \hline
 Tree Type & Env & Hyperparameter &  Value \\ \hline 
  \multirow{ 6}{*}{Common} & \multirow{ 3}{*}{CartPole-v1}  & learning rate & $1\times 10^{-3}$\\ \cline{3-4}
 & & batch size & 1280\\ \cline{3-4}
 & & epochs & 80 \\ \cline{2-4}

 & \multirow{ 3}{*}{LunarLander-v2}  & learning rate & $1\times 10^{-3}$\\ \cline{3-4}
 & & batch size & 1280\\ \cline{3-4}
 & & epochs & 80 \\ \hline \hline
 
 \multirow{ 2}{*}{SDT} & CartPole-v1  & depth & 3\\ \cline{2-4}
 & \multirow{ 1}{*}{LunarLander-v2}  & depth & 4\\ \hline
 
 \multirow{ 6}{*}{CDT} &  \multirow{ 3}{*}{CartPole-v1} & FL depth  & 2 \\ \cline{3-4}
 & & DM depth  & 2 \\ \cline{3-4}
 & & \# intermediate variables  & 2 \\ \cline{2-4}
  &  \multirow{ 3}{*}{LunarLander-v2}   & FL depth  & 3 \\ \cline{3-4}
 & & DM depth  & 3 \\ \cline{3-4}
 & & \# intermediate variables  & 2 \\
 \hline
\end{tabular}
\caption{Imitation learning hyperparameters. The "Common" hyperparameters are shared for both SDT and CDT.}
\label{tab:rl_params}
\end{table}

\section{Additional Imitation Learning Results for Stability Analysis}
\label{app:additional_il}

% stability and fidelity for reliable interpretations
Both the fidelity and stability of mimic models reflect the reliability of them as interpretable models. Fidelity is the accuracy of the mimic model, \emph{w.r.t.} the original model. It is an estimation of similarity between the mimic model and the original one in terms of prediction results. However, fidelity is not sufficient for reliable interpretations.  An unstable family of mimic models will lead to inconsistent explanations of original black-box models. The stability of the mimic model is a deeper excavation into the model itself and comparisons among several runs. Previous research~\citep{bastani2017interpreting} has investigated the fidelity and stability of decision trees as mimic models, where the stability is estimated with the fraction of equivalent nodes in different random decision trees trained under the same settings. However, in our tests, apart from evaluating the tree weights in different imitators, we also use the feature importance given by different differentiable DT instances with the same architecture and training setting to measure the stability.

\subsection{Feature Importance Assignment on Trees}
\label{app:feature_importance}

For differentiable DT methods (e.g. CDT and SDT), since the decision boundaries within each node are linear combinations of features, we can simply take the weight vector $\boldsymbol{w}^j_i$ as the importance assignment for those features within each node. 

After training the tree, a \textit{local explanation} is straightforward to derive with the inference process of a single instance and the decision path on the tree. A \textit{global explanation} can be the average local explanation across instances, \emph{e.g.} in an episode or several episodes under the RL settings.  Here we list several ways of assigning importance values for input features with SDT, to derive the feature importance vector $\boldsymbol{I}$ with the same dimension as the decision node vectors $\boldsymbol{w}$ and input feature:

For \textit{local explanation}:
\begin{itemize}
    \item \RomanNumeralCaps{1}. A trivial way of feature importance assignment on SDT would be simply adding up all weight vectors of nodes on the decision path:
    $\boldsymbol{I}(x) = \sum_{i,j}\boldsymbol{w}^j_i(\boldsymbol{x})$
    
    \item \RomanNumeralCaps{2}. The second way is a weighted average of the decision vectors, \emph{w.r.t.} the confidence of the decision boundaries for a specific instance. Considering the soft decision boundary on each node, we assume that the more confident the boundary is applied to partition the data point into a specific region within the space, the more reliable we can assign feature importance according to the boundary.
    The \textit{confidence} of a decision boundary can be positively correlated with the distance from the data point to the boundary, or the probability of the data point falling into one side of the boundary. The latter one is straightforward in our settings. We define the confidence as $p(x)=p^{\lfloor j/2 \rfloor \rightarrow j}_{i-1\rightarrow i}(x)$, which is also the probability of choosing node $j$ in $i$-th layer from its parent on instance $x$'s decision path. It indicates how far the data point is from the middle of the soft boundary in a probabilistic view.
    Therefore the importance value is derived via multiplying the confidence value with each decision node vector:
    $\boldsymbol{I}(\boldsymbol{x}) = \sum_{i,j}p^{\lfloor j/2 \rfloor \rightarrow j}_{i-1\rightarrow i}(\boldsymbol{x}) \boldsymbol{w}^j_i(\boldsymbol{x})$.

    Fig.~\ref{fig:boundary2} helps to demonstrate the reason for using the decision confidence (\emph{i.e.}, probability) as a weight for assigning feature importance, which indicates that the probability of belonging to one category is positively correlated with the distance from the instance to the decision boundary. Therefore when there are multiple boundaries for partitioning the space (\emph{e.g.}, two in the figure), we assign the boundaries having a shorter distance to the data point with smaller confidence in determining feature importance, since based on the closer boundaries the data point is much easier to be perturbed into the contrary category and less confident to remain in the original.

    \begin{figure}[H]
    \begin{center}
        \includegraphics[scale=0.25]{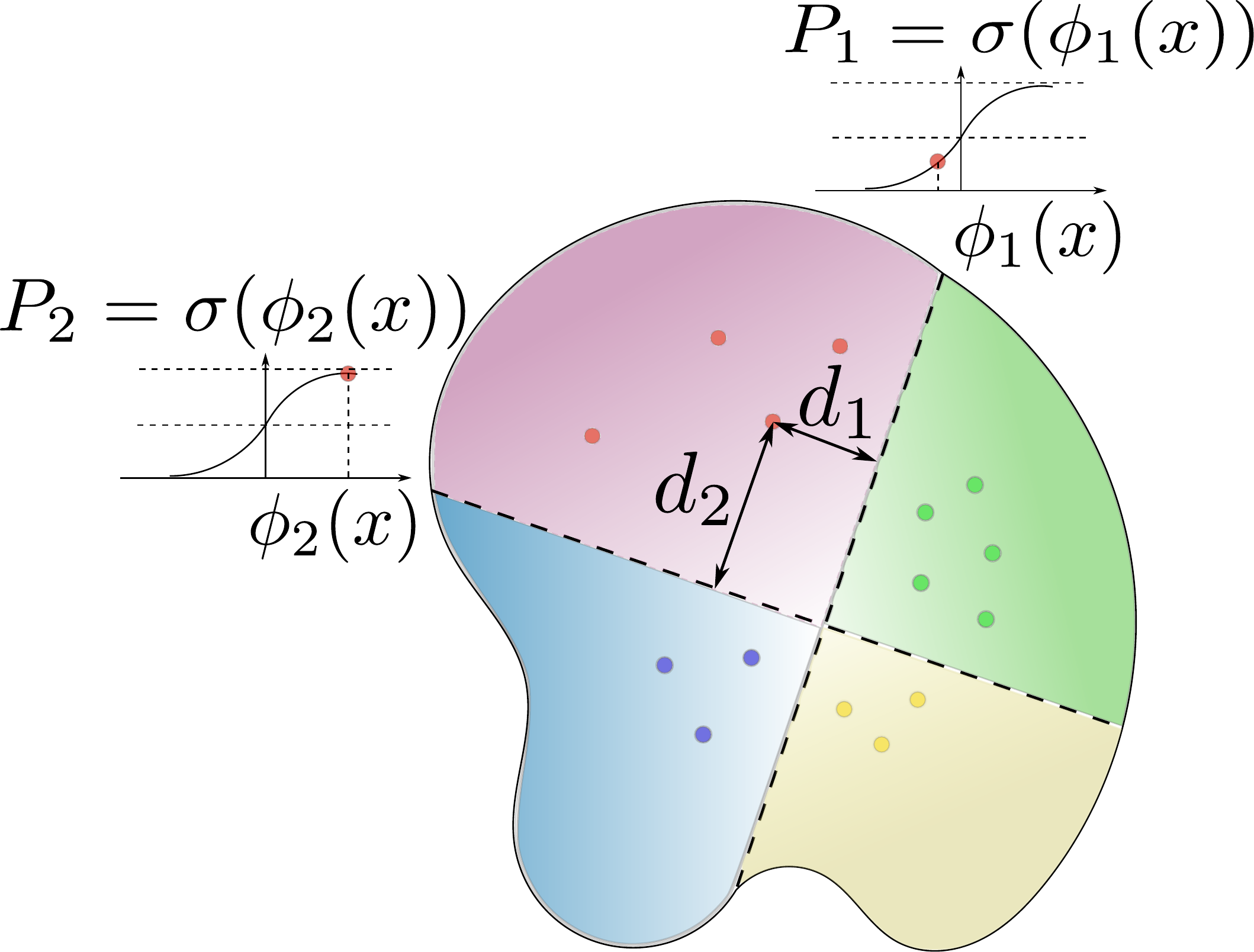}
    \end{center}
    \caption{Multiple soft decision boundaries (dashed lines) partition the space. The dots represent input data points, and different colored regions indicate different partitions in the input space. The boundaries closer to the instance are less important in determining the feature importance since they are less distinctive for the instance.}
    \label{fig:boundary2}
    \end{figure}
    
    \item \RomanNumeralCaps{3}. Since the tree we use is differentiable, we can also apply gradient-based methods for feature importance assignment, which is: $\boldsymbol{I}(\boldsymbol{x}) = \frac{\partial{y}}{\partial{\boldsymbol{x}}}$, where $ y=\text{SDT}(\boldsymbol{x})$.

\end{itemize}

For \textit{global explanation}:
\begin{itemize}
\item We can simply average the feature importance at each time step (\emph{i.e.}, local explanation) to get global feature importance over an episode or across episodes, where the local explanations can be derived in either of the above ways.
\end{itemize}

\subsection{Results of Feature Importance in Imitation Learning}
To testify the stability of applying SDT method with imitation learning from a given agent, we compare the SDT agents of different runs and original agents using certain metrics. The agent to be imitated from is a heuristic decision tree (HDT) agent, and the metric for evaluation is the assigned feature importance across an episode on each feature dimension. As described in the previous section, the feature importance for local explanation can be achieved in three ways, which work for both HDT and SDT here. The environment is \textit{LunarLander-v2} with an 8-dimensional observation in our experiments here.

Considering SDT of different runs may predict different actions, even if they are trained with the same setting and for a considerable time to achieve similarly high accuracies, we conduct comparisons not only for an online decision process during one episode, but also on a pre-collected offline state dataset by the HDT agent. We hope this can alleviate the accumulating differences in trajectories caused by consecutively different actions made by different agents, and give a more fair comparison on the decision process (or feature importance) for the same trajectory.

\begin{center}
\begin{figure}[htbp]
\includegraphics[scale=0.28]{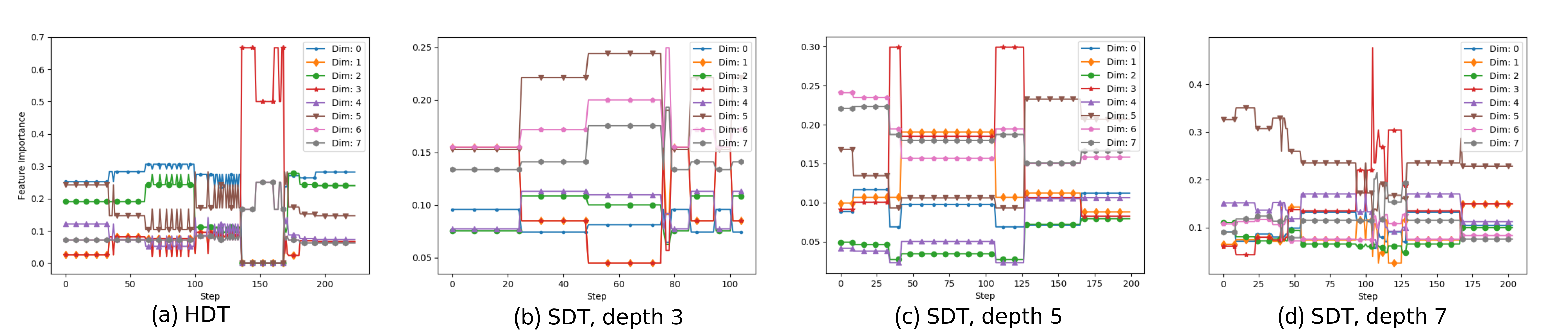}
\caption{Comparison of feature importance (local explanation \RomanNumeralCaps{1}) for SDT of depth 3, 5, 7 with HDT on an episodic decision making process.}
\label{fig:stability_online}
\end{figure}
\end{center}

\textbf{Different Tree Depths.} First, the comparison of feature importance (adding up node weights on decision path) for HDT and the learned SDT of different depths in an online decision episode is shown as Fig.~\ref{fig:stability_online}. All SDT agents are trained for 40 epochs to convergence. The accuracies of three trees are $87.35\%, 95.23\%, 97.50\%$, respectively.

% \begin{figure}
% \subfloat[HDT]{\includegraphics[width = 1.5in]{img/heuristic_tree_importance_online.png}} 
% \subfloat[SDT, depth 3]{\includegraphics[width = 1.5in]{img/sdt_importance_online_depth3.png}}
% \subfloat[SDT, depth 5]{\includegraphics[width = 1.5in]{img/sdt_importance_online_depth5.png}}
% \subfloat[SDT, depth 7]{\includegraphics[width = 1.5in]{img/sdt_importance_online_depth7.png}} 
% \caption{Comparison of feature importance (local explanation \RomanNumeralCaps{1}) for SDT of depth 3, 5, 7 with HDT on the an episodic decision making process with the same random seed.}
% \label{fig:stability_online}
% \end{figure}

From Fig.~\ref{fig:stability_online} we can tell significant differences among SDTs with different depths, as well as in comparing them against the HDT even on the episode with the same random seed, which indicates that the depth of SDT will not only affect the model prediction accuracy but also the decision making process.

\begin{figure}[H]
\setlength{\tempwidth}{.3\linewidth}
\settoheight{\tempheight}{\includegraphics[width=\tempwidth]{example-image-a}}%
\centering
\hspace{\baselineskip}
\columnname{SDT, id=1}\hfil
\columnname{SDT, id=2}\hfil
\columnname{SDT, id=3}\\
\rowname{\textit{local explanation} I}
\subfloat[]{\includegraphics[width = 1.7in]{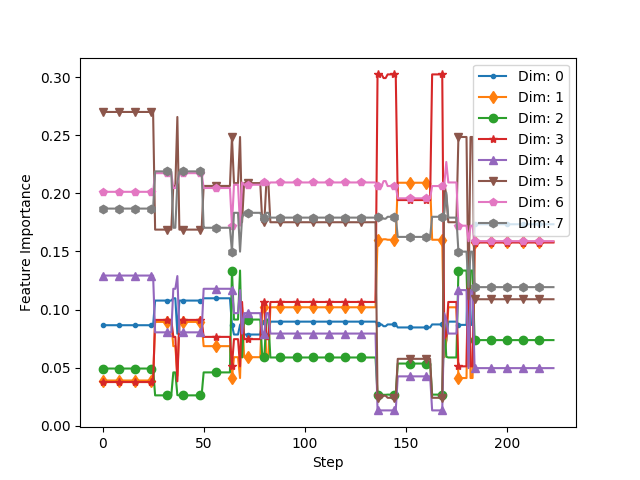}}\label{a}\hfil
\subfloat[]{\includegraphics[width = 1.7in]{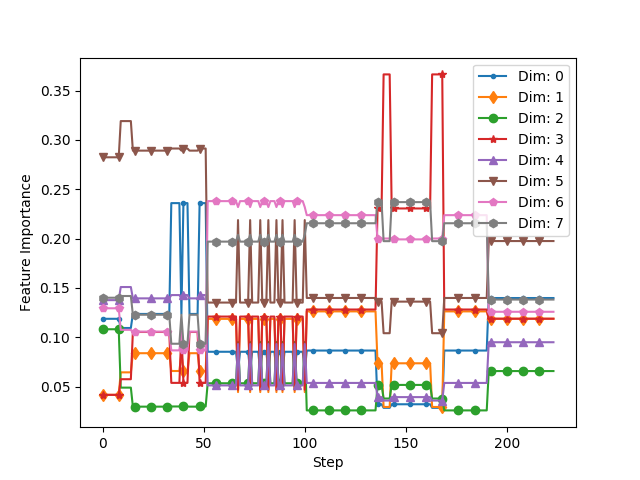}}\label{b}\hfil
\subfloat[]{\includegraphics[width = 1.7in]{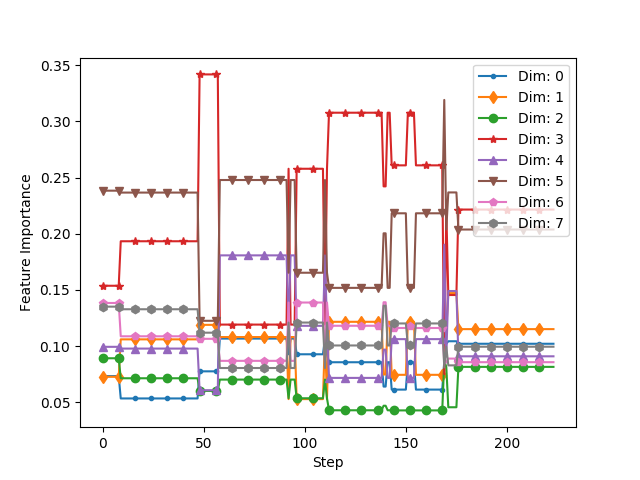}}\label{c}\\
\rowname{\textit{local explanation} II}
\subfloat[]{\includegraphics[width = 1.7in]{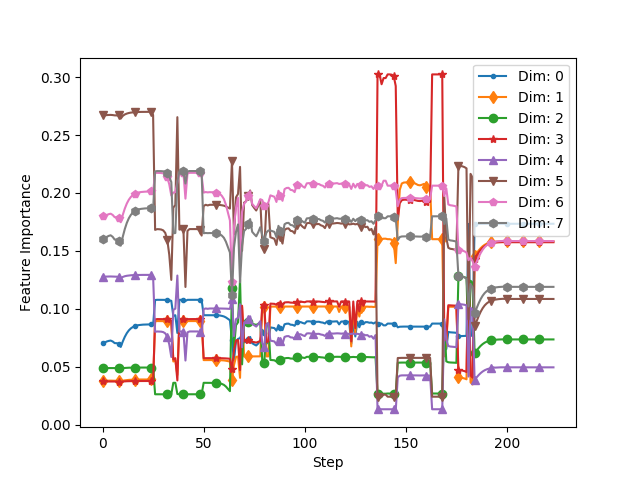}}\label{d}\hfil
\subfloat[]{\includegraphics[width = 1.7in]{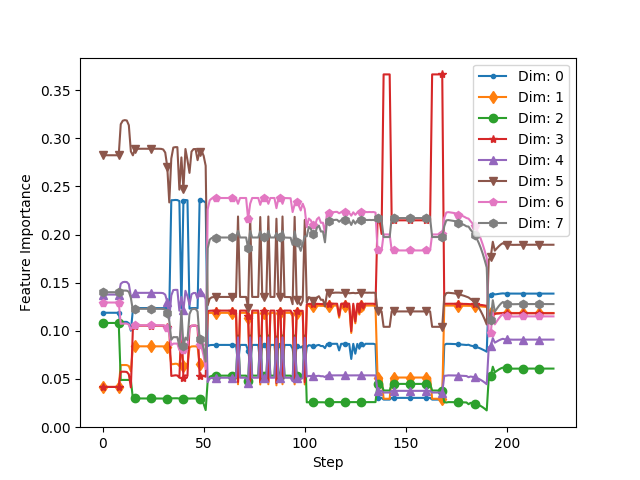}}\label{e}\hfil
\subfloat[]{\includegraphics[width = 1.7in]{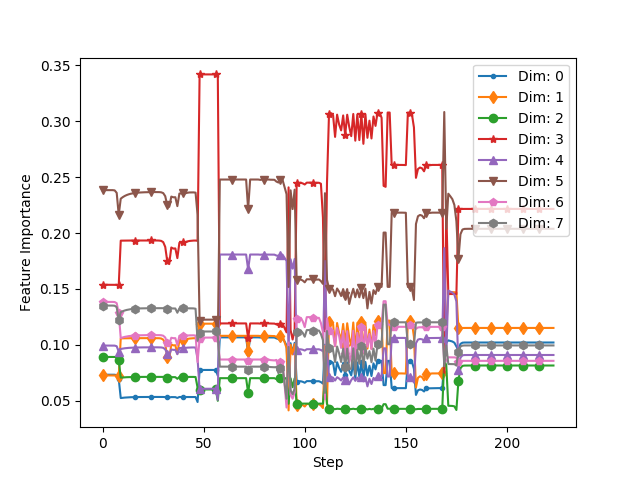}}\label{f}\\
\rowname{\textit{local explanation} III}
\subfloat[]{\includegraphics[width = 1.7in]{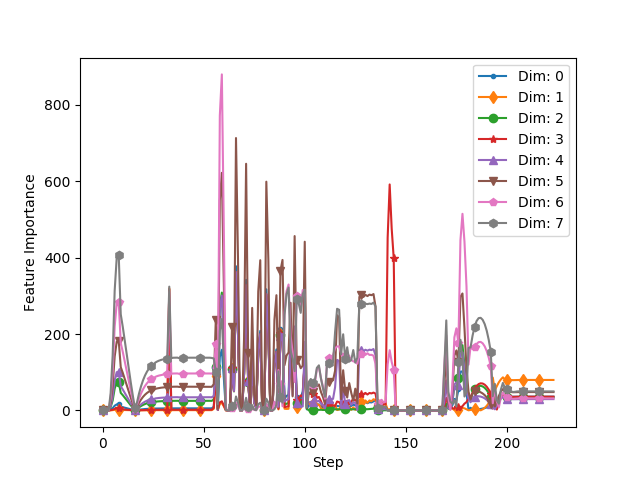}}\label{g}\hfil
\subfloat[]{\includegraphics[width = 1.7in]{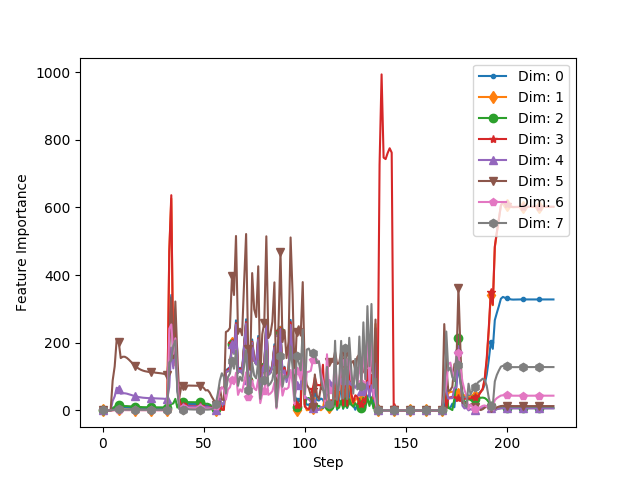}}\label{h}\hfil
\subfloat[]{\includegraphics[width = 1.7in]{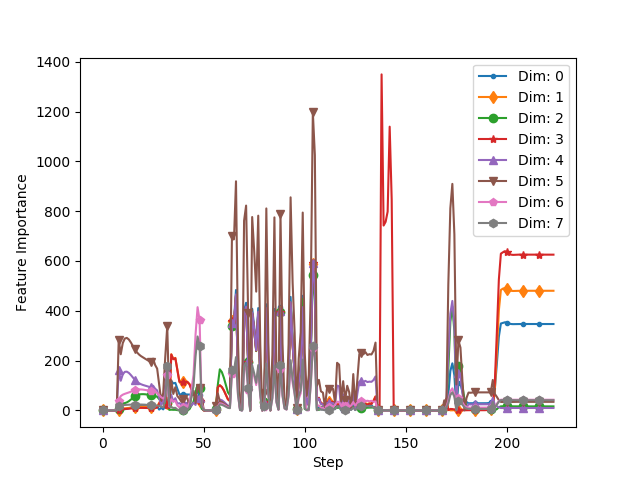}}\label{i}
\caption{Comparison of feature importance for three SDTs (depth=5, trained under the same setting) with three different local explanations. All runs are conducted on the same offline episode.}
\label{fig:feature_importance_compare}
\end{figure}

\textbf{Same Tree with Different Runs.} We compare the feature importance on an offline dataset, containing the states of the HDT agent encounters in one episode. All SDT agents have a depth of 5 and are trained for 80 epochs to convergence. The three agents have testing accuracies of $95.88\%, 97.93\%, \text{ and } 97.79\%$ respectively after training. 
% A SDT imitating a pre-trained PPO policy is also involved, which is less well-behaved compared with the other three directly imitating the heuristic agent's behaviors. 
The feature importance values are evaluated with different approaches as mentioned above (\textit{local explanation} \RomanNumeralCaps{1}, \RomanNumeralCaps{2} and \RomanNumeralCaps{3}) on the same offline episode, as shown in Fig~\ref{fig:feature_importance_compare}. In the results, \textit{local explanation} II and III looks similar, since most decision nodes in the decision path with greatest probability have the probability values close to 1 (\emph{i.e.} close to a hard decision boundary) when going to the child nodes.

From Fig.~\ref{fig:feature_importance_compare}, considerable differences can also be spotted in different runs for local explanations, even though the SDTs have similar prediction accuracies, no matter which metric is applied.

% \begin{figure}[h]
% \centering
% % \subfloat[HDT]{\includegraphics[width = 1.5in]{img/heuristic_tree_importance_offline.png}} 
% % \subfloat[SDT, PPO]{\includegraphics[width = 1.5in]{img/sdt_importance_offline_ppo.png}} 
% \rowname{Exp 1}
% \subfloat[SDT, id=1]{\includegraphics[width = 2.in]{img/sdt_importance_offline4.png}}
% \subfloat[SDT, id=2]{\includegraphics[width = 2.in]{img/sdt_importance_offline5.png}}
% \subfloat[SDT, id=3]{\includegraphics[width = 2.in]{img/sdt_importance_offline6.png}} 

% \subfloat[SDT, id=1]{\includegraphics[width = 2.in]{img/sdt_importance_offline4_weighted.png}}
% \subfloat[SDT, id=2]{\includegraphics[width = 2.in]{img/sdt_importance_offline5_weighted.png}}
% \subfloat[SDT, id=3]{\includegraphics[width = 2.in]{img/sdt_importance_offline6_weighted.png}} 

% \subfloat[SDT, id=1]{\includegraphics[width = 2.in]{img/sdt_importance_offline_grad1.png}}
% \subfloat[SDT, id=2]{\includegraphics[width = 2.in]{img/sdt_importance_offline_grad2.png}}
% \subfloat[SDT, id=3]{\includegraphics[width = 2.in]{img/sdt_importance_offline_grad3.png}}
% \caption{Comparison of feature importance for SDT (depth=5). All runs are conducted on the same offline episode. }
% \label{fig:stability_offline_weight}
% \end{figure}

% \begin{figure}[H]
% \centering
% % \subfloat[HDT]{\includegraphics[width = 1.5in]{img/heuristic_tree_importance_offline.png}} 
% % \subfloat[SDT, PPO]{\includegraphics[width = 1.5in]{img/sdt_importance_offline_ppo_weighted.png}} 
% \subfloat[SDT, id=1]{\includegraphics[width = 2.in]{img/sdt_importance_offline4_weighted.png}}
% \subfloat[SDT, id=2]{\includegraphics[width = 2.in]{img/sdt_importance_offline5_weighted.png}}
% \subfloat[SDT, id=3]{\includegraphics[width = 2.in]{img/sdt_importance_offline6_weighted.png}} 
% \caption{Comparison of feature importance (local explanation \RomanNumeralCaps{2}) for SDT (depth=5).  Three runs are conducted on an offline episode.}
% \label{fig:stability_offline_weight_probs}
% \end{figure}

% \begin{figure}[H]
% \centering
% % \subfloat[SDT, PPO]{\includegraphics[width = 1.5in]{img/sdt_importance_offline_grad_ppo.png}} 
% \subfloat[SDT, id=1]{\includegraphics[width = 2.in]{img/sdt_importance_offline_grad1.png}}
% \subfloat[SDT, id=2]{\includegraphics[width = 2.in]{img/sdt_importance_offline_grad2.png}}
% \subfloat[SDT, id=3]{\includegraphics[width = 2.in]{img/sdt_importance_offline_grad3.png}}
% \caption{Comparison of feature importance (local explanation \RomanNumeralCaps{3}) for SDT (depth=5). Three runs are conducted on an offline episode.}
% \label{fig:stability_offline_grad}
% \end{figure}

\subsection{Tree Structures in Imitation Learning}
\label{app:tree_structures}
\begin{figure}[H]
    \begin{center}
        \includegraphics[scale=0.21]{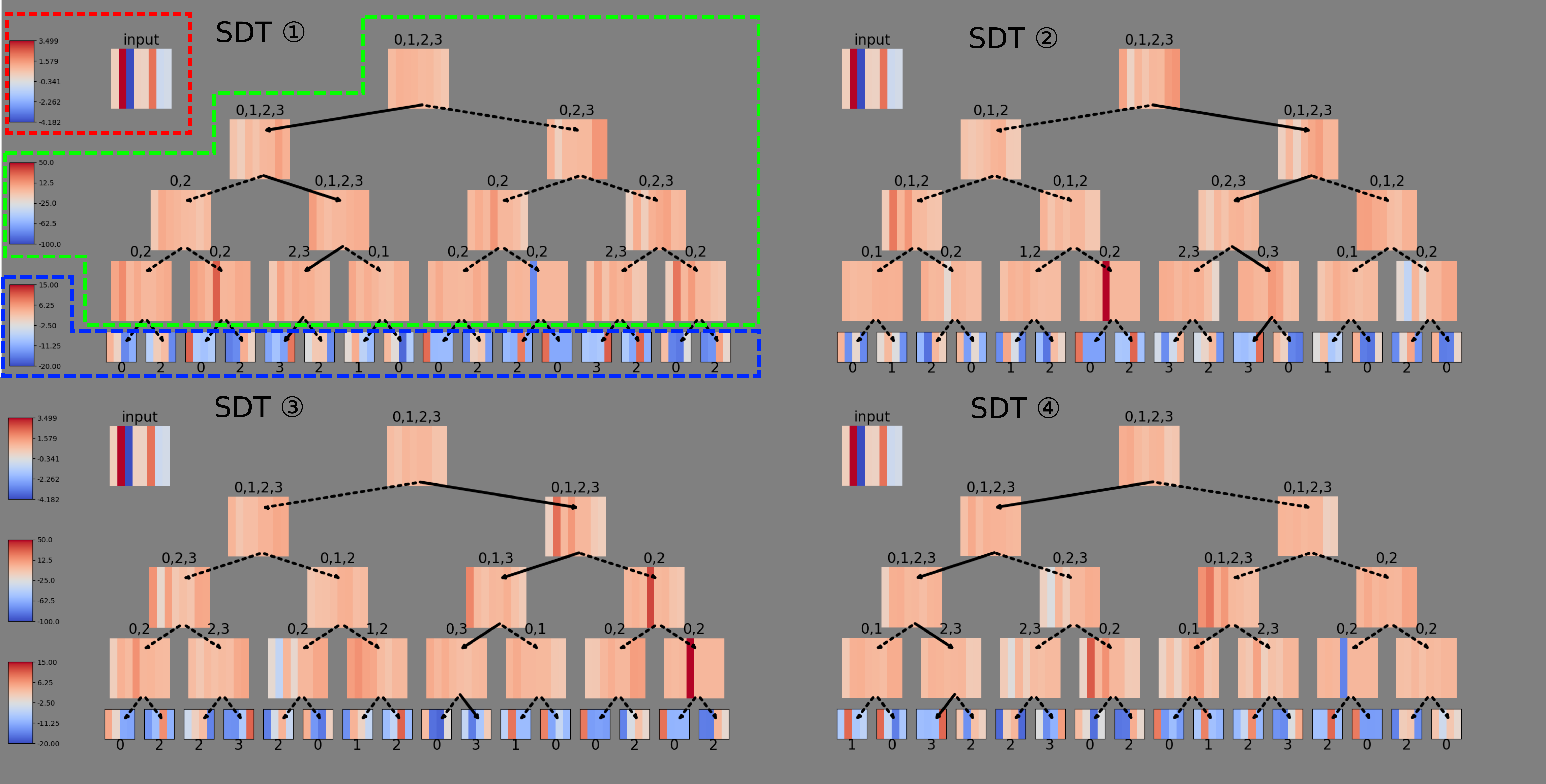}
    \end{center}
    \caption{Comparison of four runs with the same setting for SDT (before discretization) imitation learning on \textit{LunarLander-v2}. The dashed lines with different colors on the left top diagram indicate the valid regions for each color bar, which is the default setting for the rest diagrams.}
    \label{fig:sdt_vis}
\end{figure}

\begin{figure}[H]
    \begin{center}
        \includegraphics[scale=0.21]{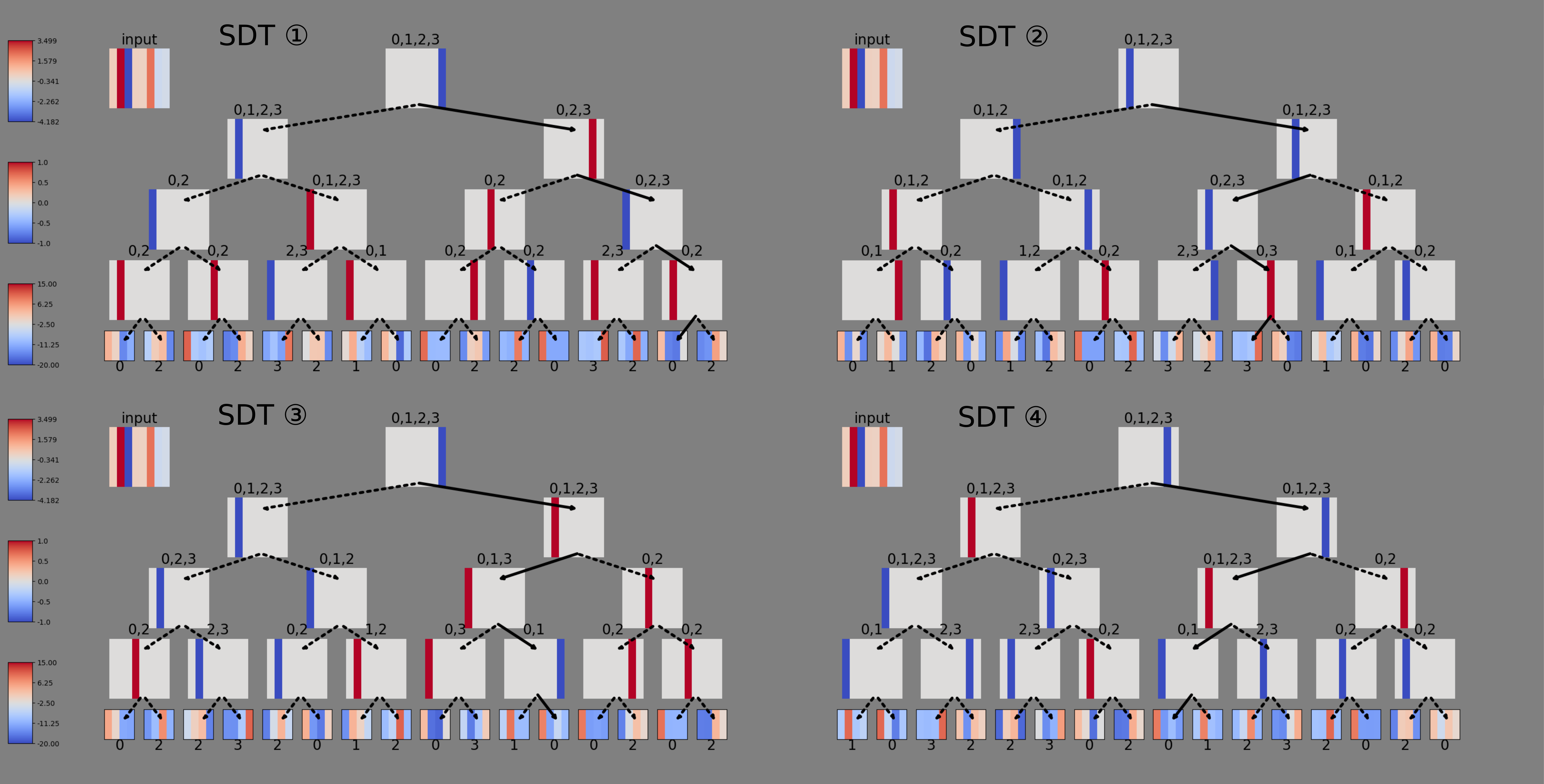}
    \end{center}
    \caption{Comparison of four runs with the same setting for SDT (after discretization) imitation learning on \textit{LunarLander-v2}.}
    \label{fig:sdt_vis_dis}
\end{figure}

\begin{figure}[H]
    \begin{center}
        \includegraphics[scale=0.4]{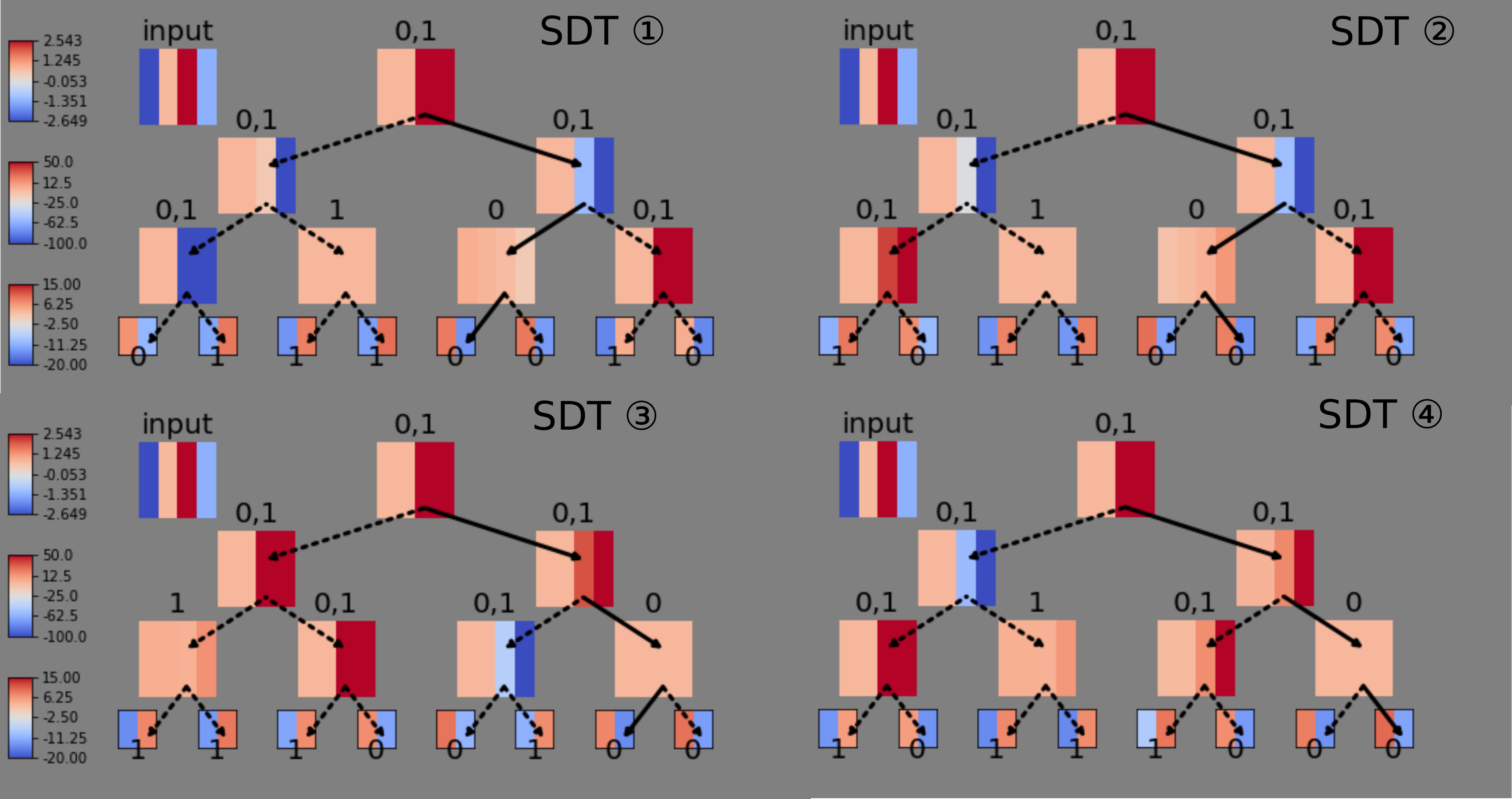}
    \end{center}
    \caption{Comparison of four runs with the same setting for SDT (before discretization) imitation learning on \textit{CartPole-v1}.}
    \label{fig:sdt_vis_cartpole}
\end{figure}

\begin{figure}[H]
    \begin{center}
        \includegraphics[scale=0.4]{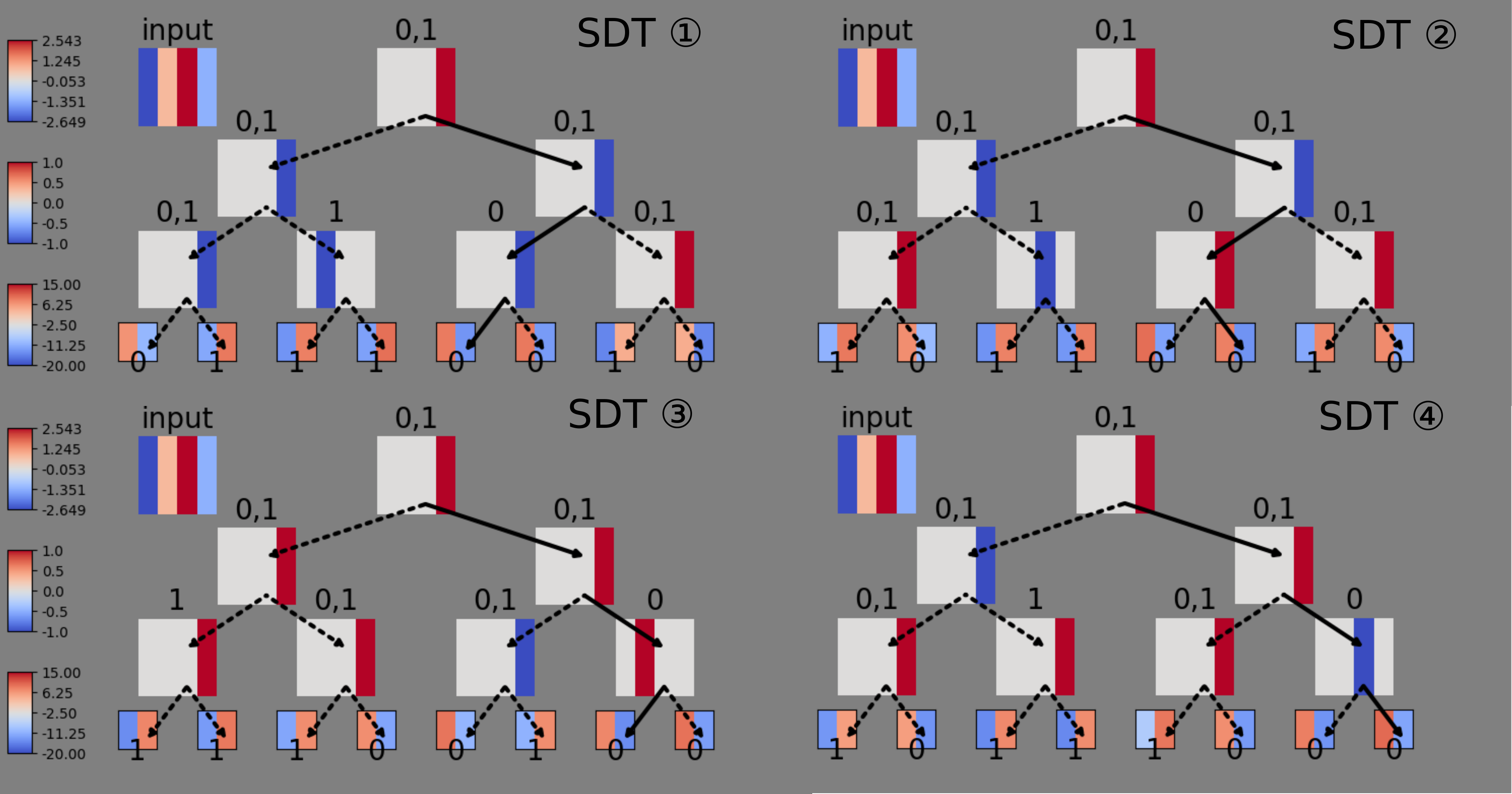}
    \end{center}
    \caption{Comparison of four runs with the same setting for SDT (after discretization) imitation learning on \textit{CartPole-v1}.}
    \label{fig:sdt_vis_dis_cartpole}
\end{figure}

\begin{figure}[H]
    \begin{center}
        \includegraphics[scale=0.24]{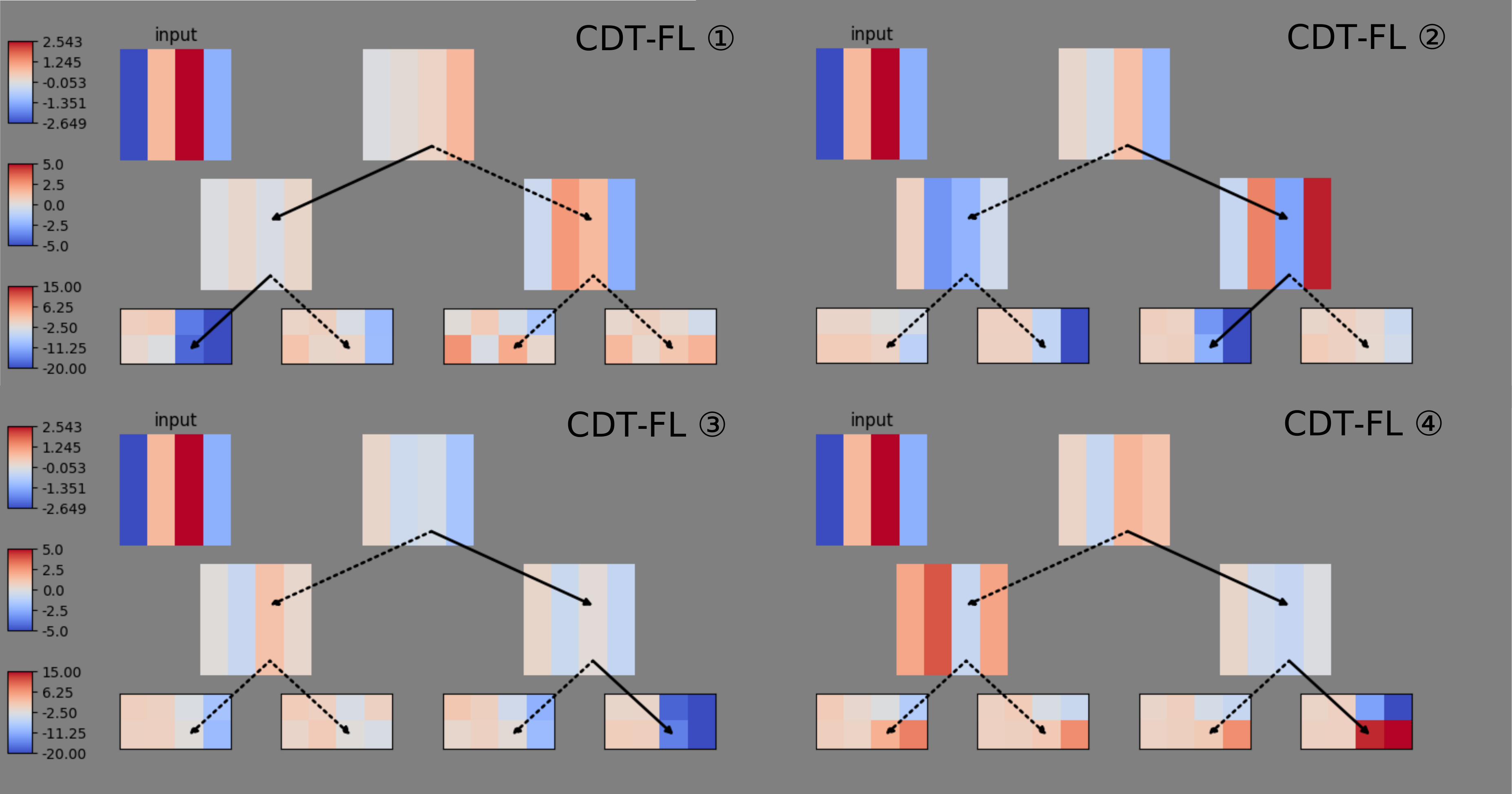} \\
        \includegraphics[scale=0.24]{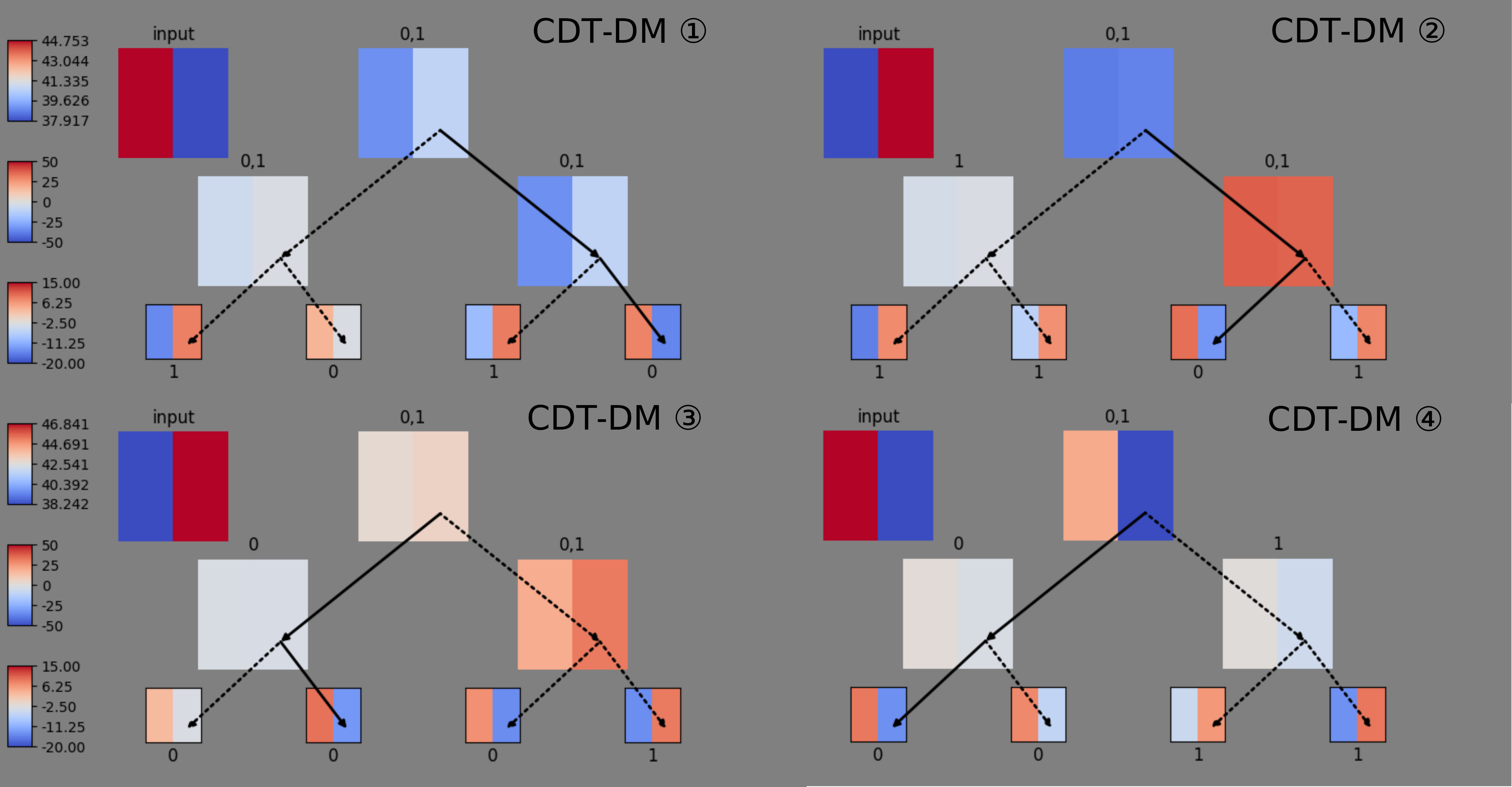}
    \end{center}
    \caption{Comparison of four runs with the same setting for CDT (before discretization) imitation learning on \textit{CartPole-v1}: feature learning trees (top) and decision making trees (bottom).}
    \label{fig:cdt_vis_cartpole}
\end{figure}

\begin{figure}[H]
    \begin{center}
        \includegraphics[scale=0.24]{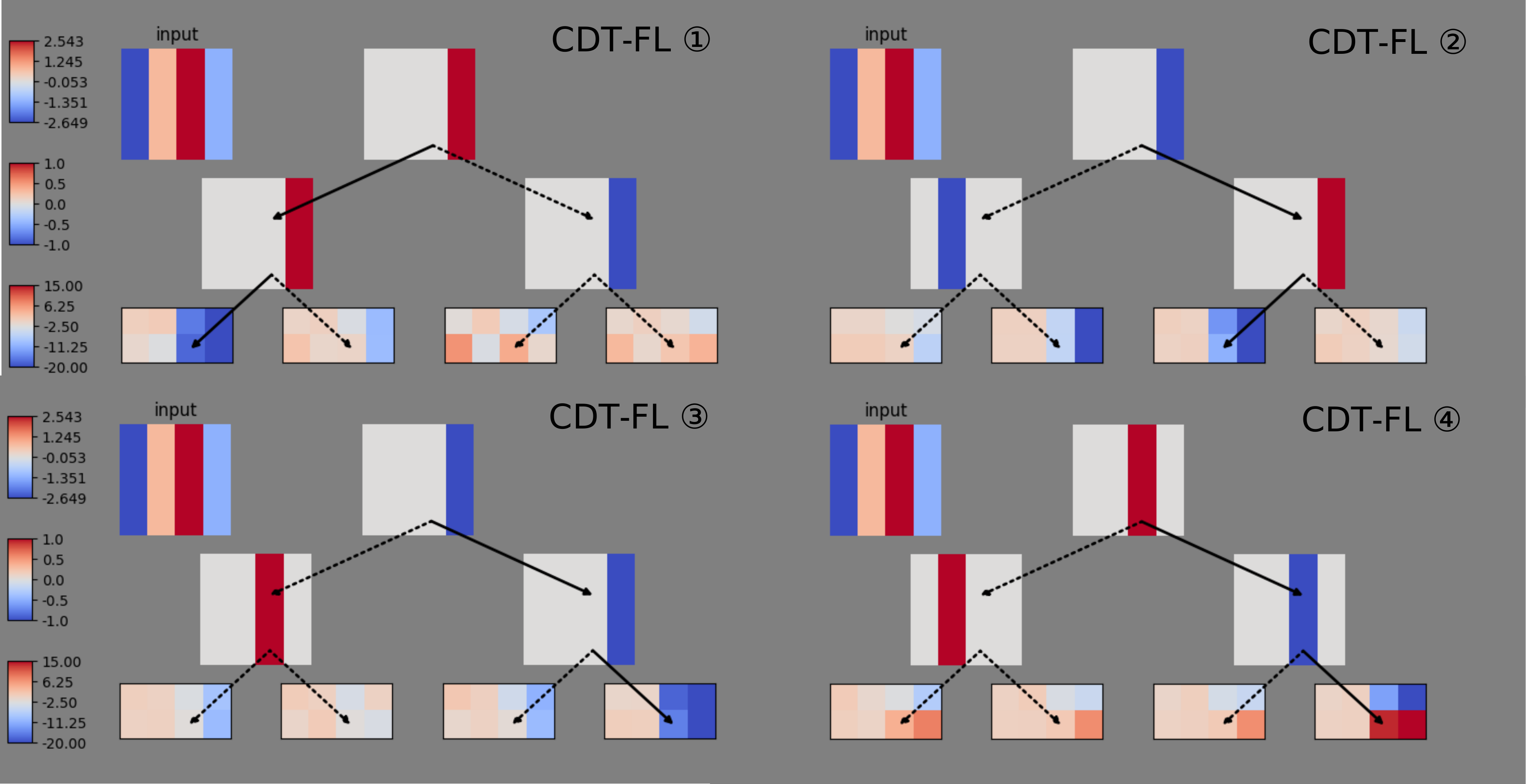} \\
        \includegraphics[scale=0.24]{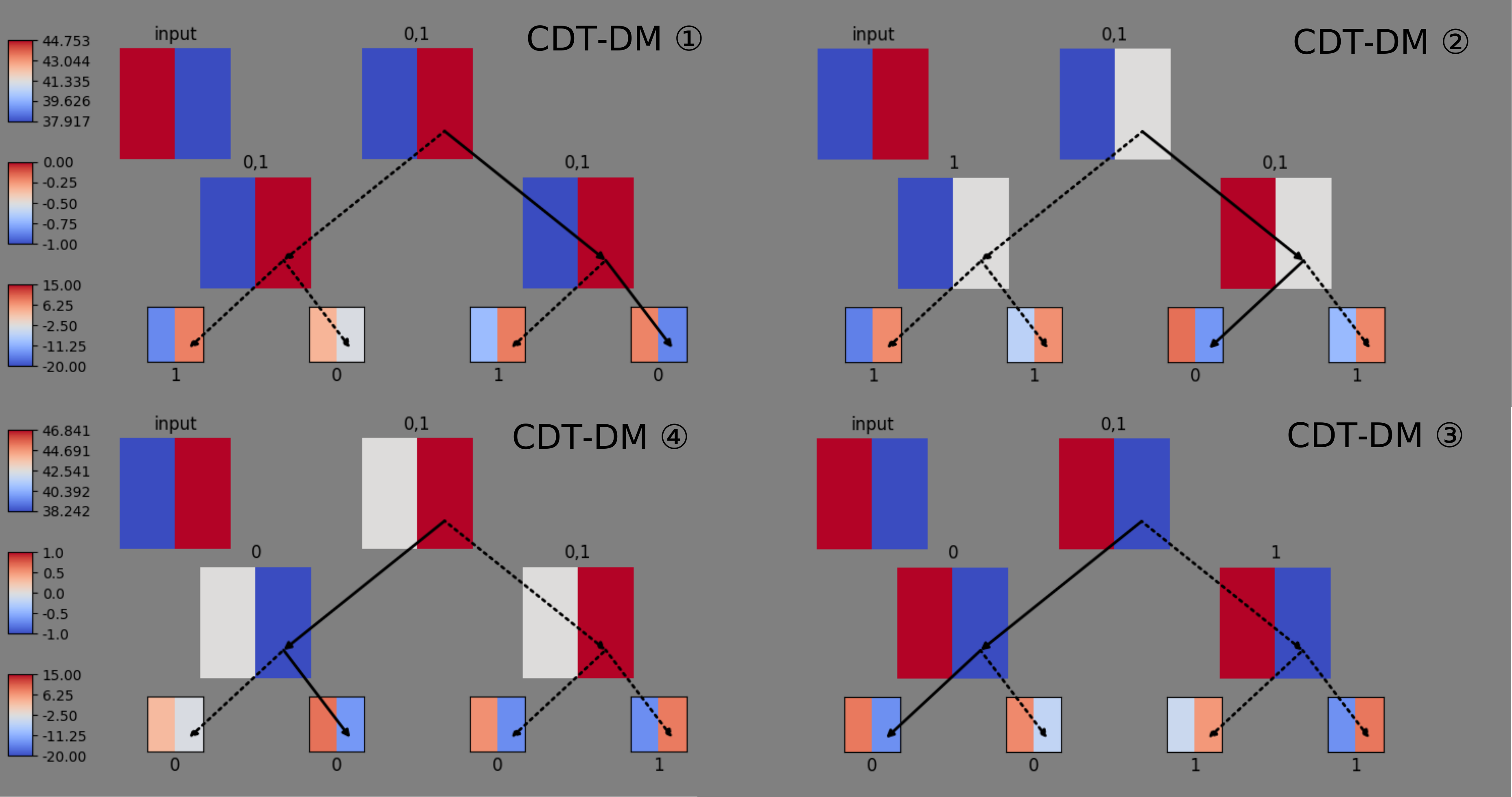}
    \end{center}
    \caption{Comparison of four runs with the same setting for CDT (after discretization) imitation learning on \textit{CartPole-v1}: feature learning trees (top) and decision making trees (bottom).}
    \label{fig:cdt_vis_dis_cartpole}
\end{figure}

We display the agents trained with CDTs and SDTs on both \textit{CartPole-v1} and \textit{LunarLander-v2} before and after tree discretization in this section, as in Fig.~\ref{fig:sdt_vis}, \ref{fig:sdt_vis_dis}, \ref{fig:sdt_vis_cartpole}, \ref{fig:sdt_vis_dis_cartpole}, \ref{fig:cdt_vis}, \ref{fig:cdt_vis_cartpole}, \ref{fig:cdt_vis_dis_cartpole}. Each figure contains trees trained in four runs with the same setting. Each sub-figure contains one learned tree (plus an input example and its output) with an inference path (\emph{i.e.}, the solid lines) for the same input instance. The lines and arrows indicate the connections among tree nodes. The colors of the squares on tree nodes show the values of weight vectors for each node. For feature learning trees in CDTs, the leaf nodes are colored with the feature coefficients. The output leaf nodes of both SDTs and decision making trees in CDTs are colored with the output categorical distributions. Three color bars are displayed on the left side for inputs, tree inner nodes, and output leaves respectively, as demonstrated in Fig.~\ref{fig:sdt_vis}. It remains the same for the rest tree plots. The digits on top of each node represent the output action categories.

Among all the learned tree structures, significant differences can be told from weight vectors, as well as intermediate features in CDTs, even if the four trees are under the same training setting. 
This will lead to considerably different explanations or feature importance assignments on trees.

\begin{figure}[H]
    \begin{center}
        \includegraphics[scale=0.25]{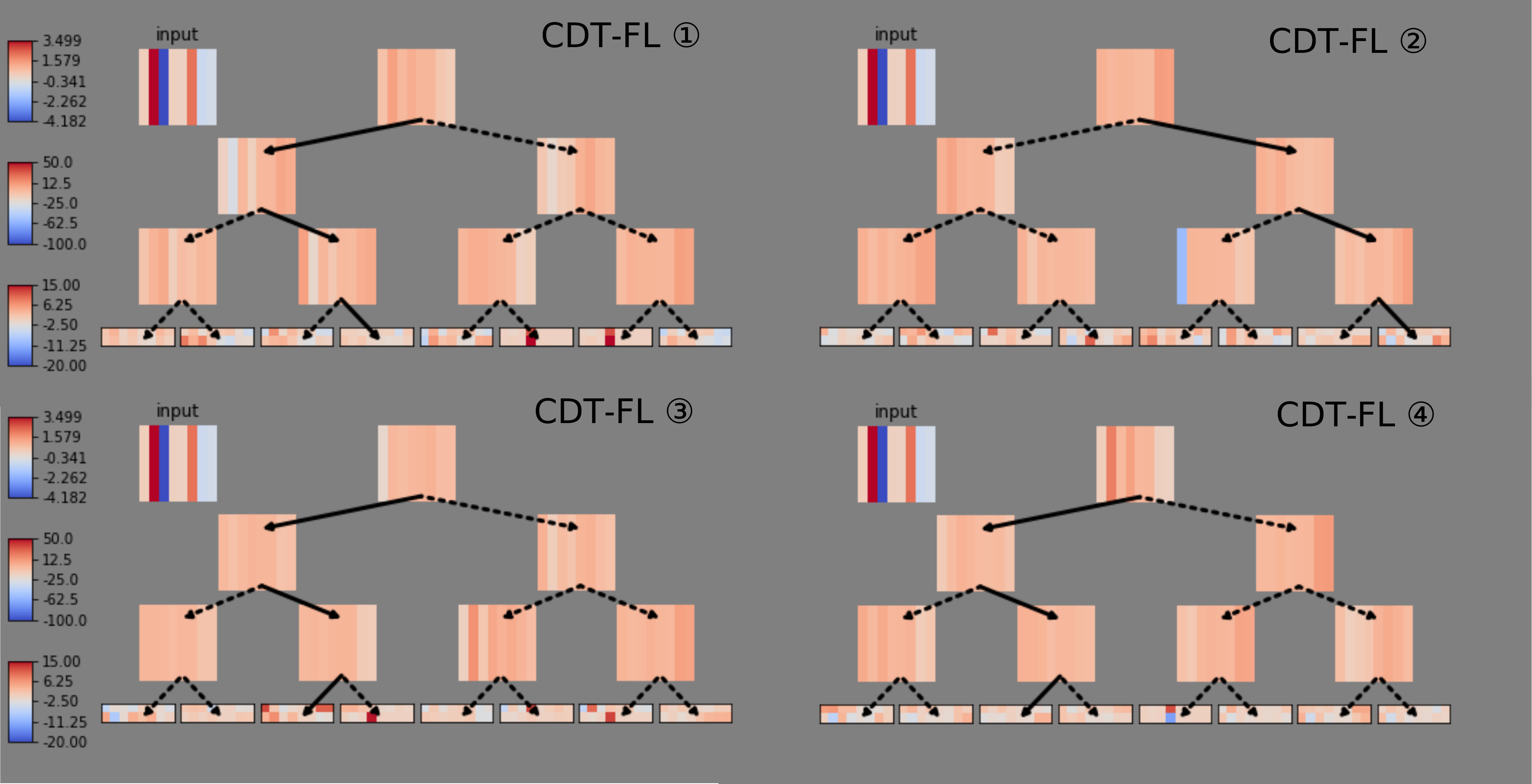} \\
        \includegraphics[scale=0.25]{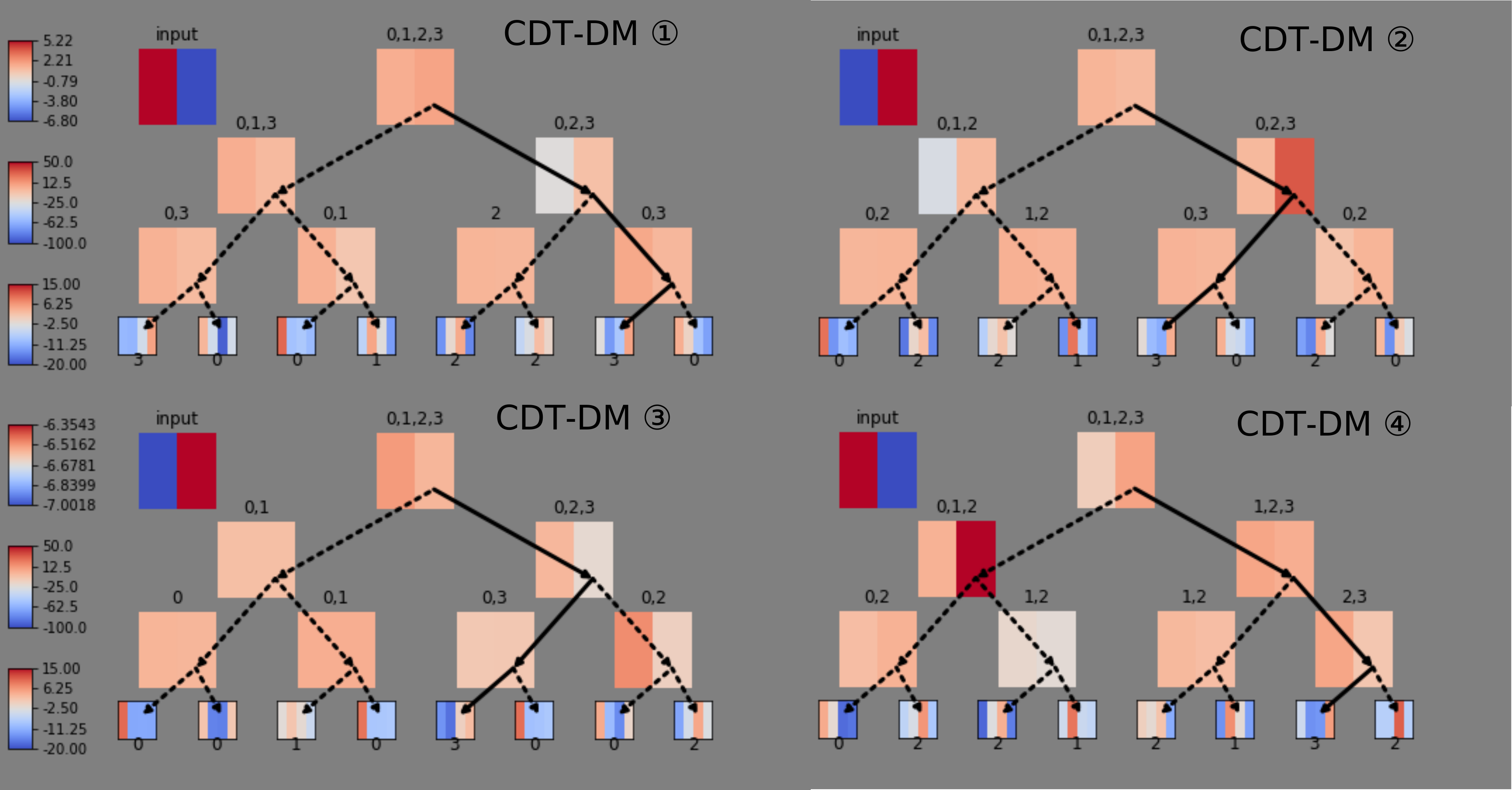}
    \end{center}
    \caption{Comparison of four runs with the same setting for CDT (before discretization) imitation learning on \textit{LunarLander-v2}: feature learning trees (top) and decision making trees (bottom).}
    \label{fig:cdt_vis}
\end{figure}

\begin{figure}[H]
    \begin{center}
        \includegraphics[scale=0.24]{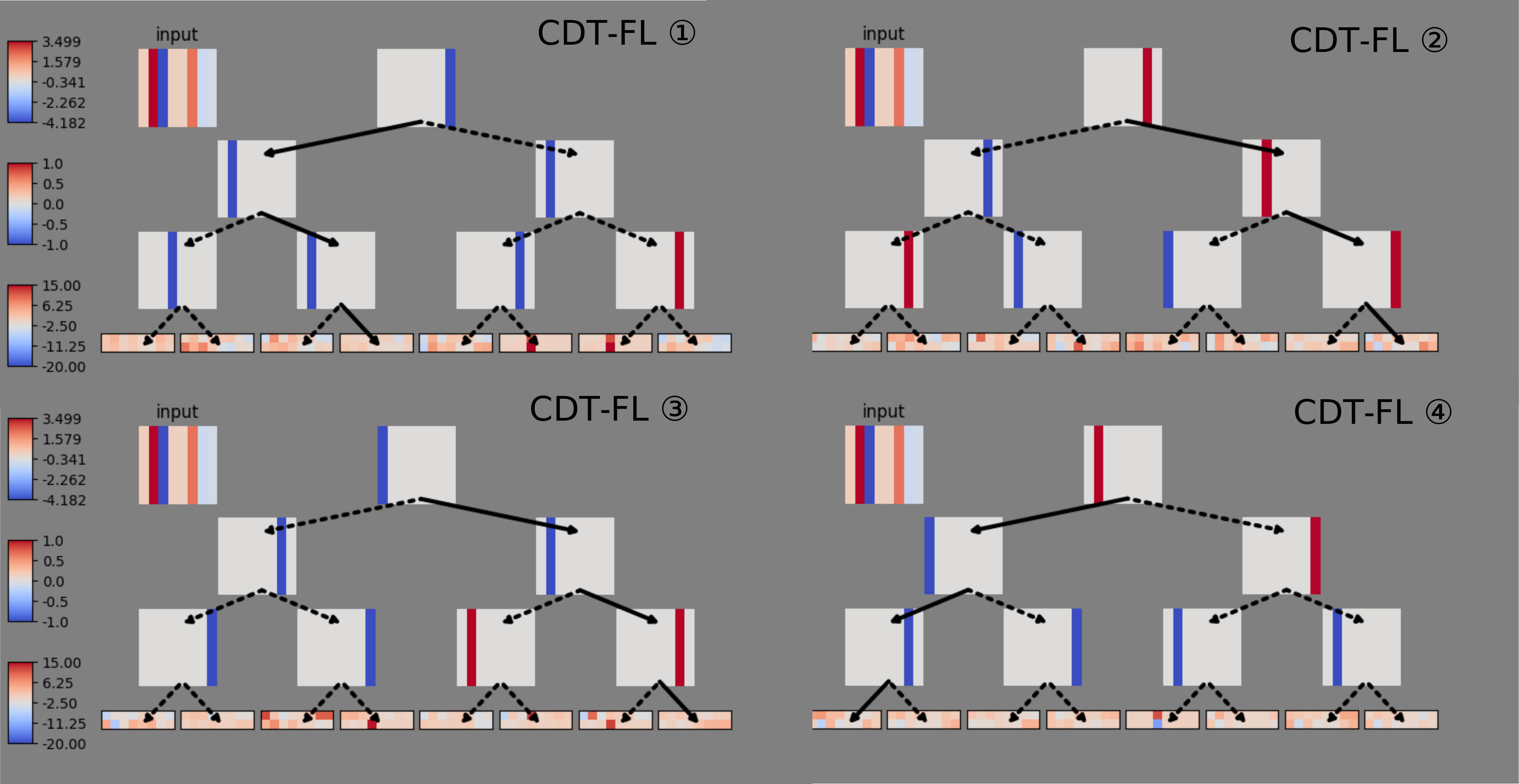} \\
        \includegraphics[scale=0.24]{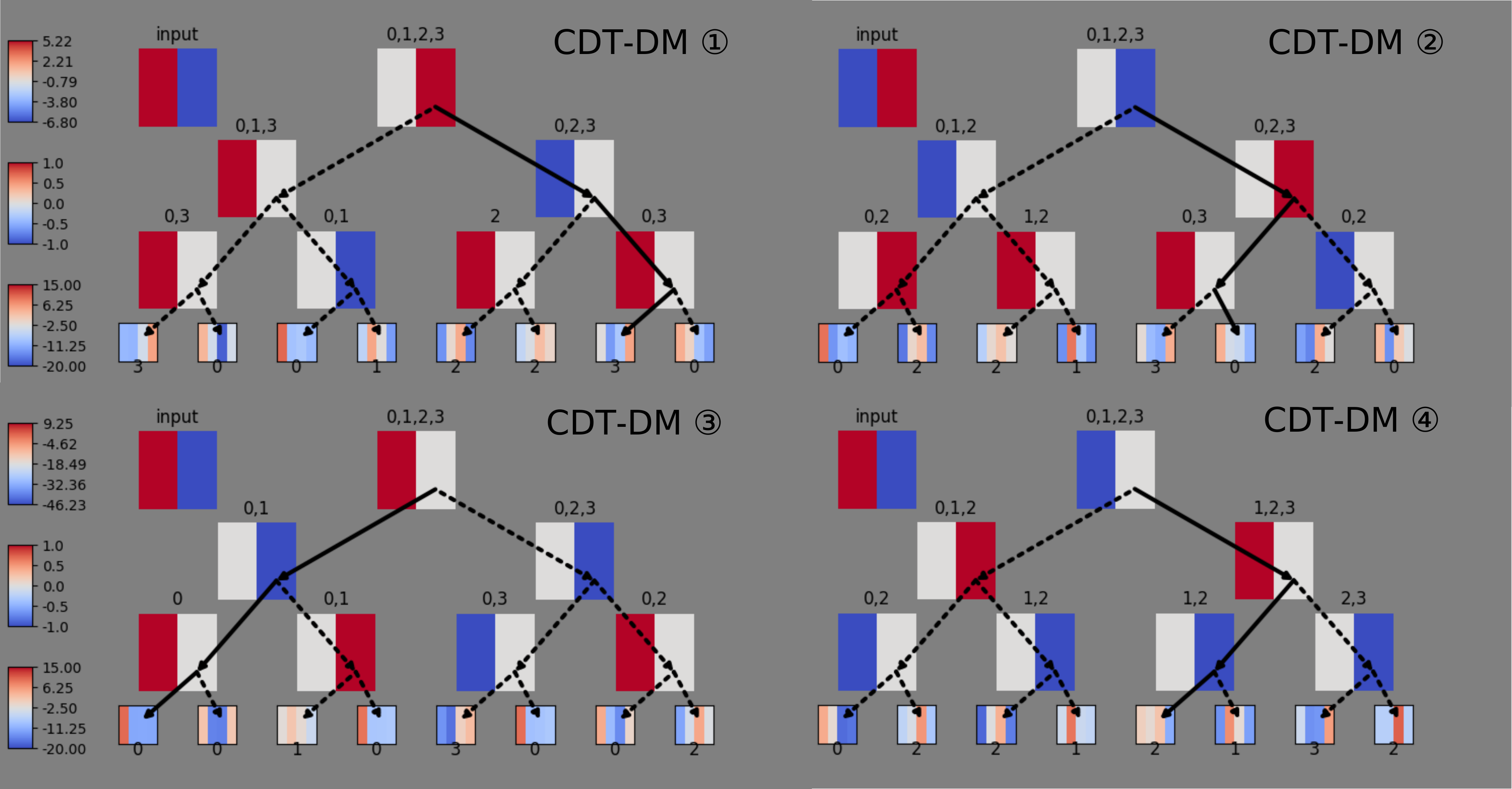}
    \end{center}
    \caption{Comparison of four runs with the same setting for CDT (after discretization) imitation learning on \textit{LunarLander-v2}: feature learning trees (top) and decision making trees (bottom).}
    \label{fig:cdt_vis}
\end{figure}

\section{Training Details in Reinforcement Learning}
\label{app:rl_params}
\begin{table}[H]
\centering
% \begin{tabular}{ |c|c|c|c|c| } 
\begin{tabular}{ m{2cm} m{3cm} m{4cm} m{2cm} } 
 \hline
 Tree Type & Env & Hyperparameter &  Value \\ \hline 
  \multirow{ 24}{*}{Common} & \multirow{ 8}{*}{CartPole-v1}  & learning rate & $5\times 10^{-4}$\\ \cline{3-4}
 & & $\gamma$ & 0.98\\ \cline{3-4}
 & & $\lambda$ & 0.95\\ \cline{3-4}
 & & $\epsilon$ & 0.1\\ \cline{3-4}
 & & update iteration & 3\\ \cline{3-4}
 & & hidden dimension (value) & 128\\ \cline{3-4}
 & & episodes & 3000 \\ \cline{3-4}
   && time horizon & 1000 \\ \cline{2-4}

 & \multirow{ 8}{*}{LunarLander-v2}  & learning rate & $5\times 10^{-4}$\\ \cline{3-4}
 & & $\gamma$ & 0.98\\ \cline{3-4}
 & & $\lambda$ & 0.95\\ \cline{3-4}
 & & $\epsilon$ & 0.1\\ \cline{3-4}
 & & update iteration & 3\\ \cline{3-4}
 & & hidden dimension (value) & 128\\ \cline{3-4}
  & & episodes & 5000 \\ \cline{3-4}
  && time horizon & 1000 \\ \cline{2-4}
  
   & \multirow{ 8}{*}{MountainCar-v0}  & learning rate & $5\times 10^{-3}$\\ \cline{3-4}
 & & $\gamma$ & 0.999\\ \cline{3-4}
 & & $\lambda$ & 0.98\\ \cline{3-4}
 & & $\epsilon$ & 0.1\\ \cline{3-4}
 & & update iteration & 10\\ \cline{3-4}
 & & hidden dimension (value) & 32\\ \cline{3-4}
  & & episodes & 5000 \\ \cline{3-4}
  && time horizon & 1000 \\ \hline \hline
  
   \multirow{ 3}{*}{MLP} & CartPole-v1  & hidden dimension (policy) & 128\\ \cline{2-4}
 & \multirow{ 1}{*}{LunarLander-v2}  & hidden dimension (policy) & 128\\ \cline{2-4}
 & \multirow{ 1}{*}{MountainCar-v0}  & hidden dimension (policy) & 32\\ \hline
 
 \multirow{ 2}{*}{SDT} & CartPole-v1  & depth & 3\\ \cline{2-4}
 & \multirow{ 1}{*}{LunarLander-v2}  & depth & 4\\ \cline{2-4}
 & \multirow{ 1}{*}{MountainCar-v0}  & depth & 3\\ \hline
 
 \multirow{ 6}{*}{CDT} &  \multirow{ 3}{*}{CartPole-v1} & FL depth  & 2 \\ \cline{3-4}
 & & DM depth  & 2 \\ \cline{3-4}
 & & \# intermediate variables  & 2 \\ \cline{2-4}
  &  \multirow{ 3}{*}{LunarLander-v2}   & FL depth  & 3 \\ \cline{3-4}
 & & DM depth  & 3 \\ \cline{3-4}
 & & \# intermediate variables  & 2 \\\cline{2-4}
   &  \multirow{ 3}{*}{MountainCar-v0}   & FL depth  & 2 \\ \cline{3-4}
 & & DM depth  & 2 \\ \cline{3-4}
 & & \# intermediate variables  & 1 \\
 \hline
\end{tabular}
\caption{RL hyperparameters. The "Common" hyperparameters are shared for both SDT and CDT.}
\label{tab:rl_params}
\end{table}

To normalize the states\footnote{We found that sometimes the state normalization can affect the learning performances significantly, especially in RL settings.  }, we collect 3000 episodes of samples for each environment with a well-trained policy and calculate its mean and standard deviation. During training, each state input is subtracted by the mean and divided by the standard deviation. 

The hyperparameters for RL are provided in Table~\ref{tab:rl_params} for MLP, SDT, and CDT on three environments.

\section{Additional Reinforcement Learning Results}
\label{app:additional_rl}
\begin{figure}[H]
    \centering
        \centering\includegraphics[scale=0.35]{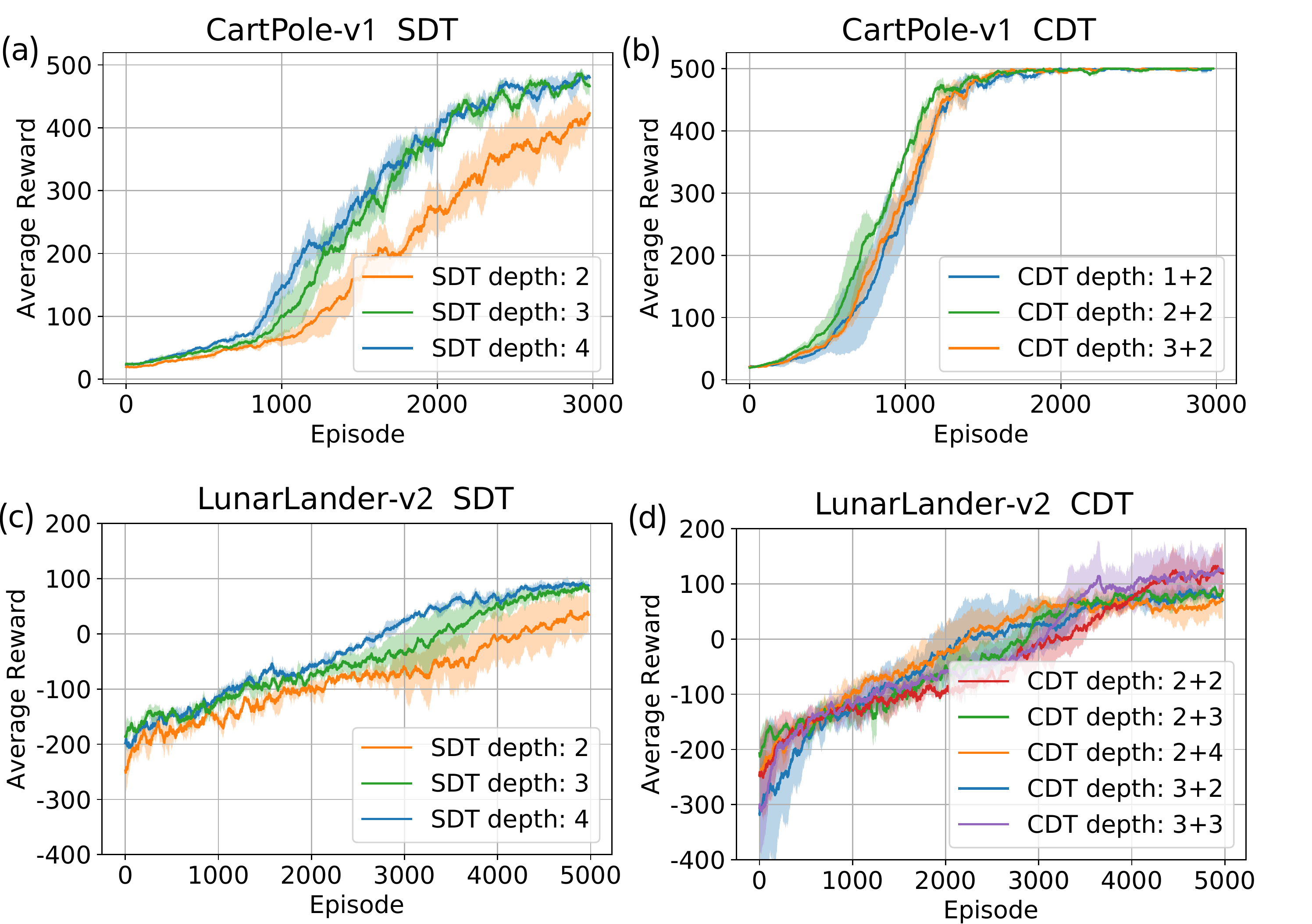}
    \caption{Comparison of SDTs and CDTs with different depths (state unnormalized). (a) and (b) are trained on \textit{CartPole-v1}, while (c) and (d) are on \textit{LunarLander-v2}.}
    \label{fig:rl_compare_depth}
\end{figure}
Fig.~\ref{fig:rl_compare_depth} displays the comparison of learning curves for SDTs and CDTs with different depths, under the RL settings without state normalization. The results are similar as those with state normalization in the main paragraph.

% \begin{figure}[H]
%     \centering
%         \centering\includegraphics[scale=0.25]{img/rl_compare_mountaincar.pdf}
%     \caption{Comparison of SDTs and CDTs on \textit{MountainCar-v0} in terms of average rewards in RL setting: (a) uses normalized input states while (b) uses unnormalized ones.}
%     \label{fig:rl_compare_mountaincar}
% \end{figure}
% Fig.~\ref{fig:rl_compare_mountaincar} shows the comparison of MLP, SDT, and CDT as policy function approximators in RL for the \textit{MountainCar-v0} environment, where the learning curves for each run, as well as their means and standard deviations, are displayed. The MLP model has two layers with 32 hidden units. The depth of SDT is 3. CDT has depths 2 and 2 for the feature learning tree and decision making tree respectively, with the dimension of the intermediate feature as 1. The training results are less stable due to large variances in exploration, but CDTs generally perform better than SDTs with near-optimal agents learned considering both cases.

\section{Trees Structures Comparison}
\label{app:tree_structure_compare}
\begin{figure}[htbp]
    \centering
        \centering
        \includegraphics[scale=0.24]{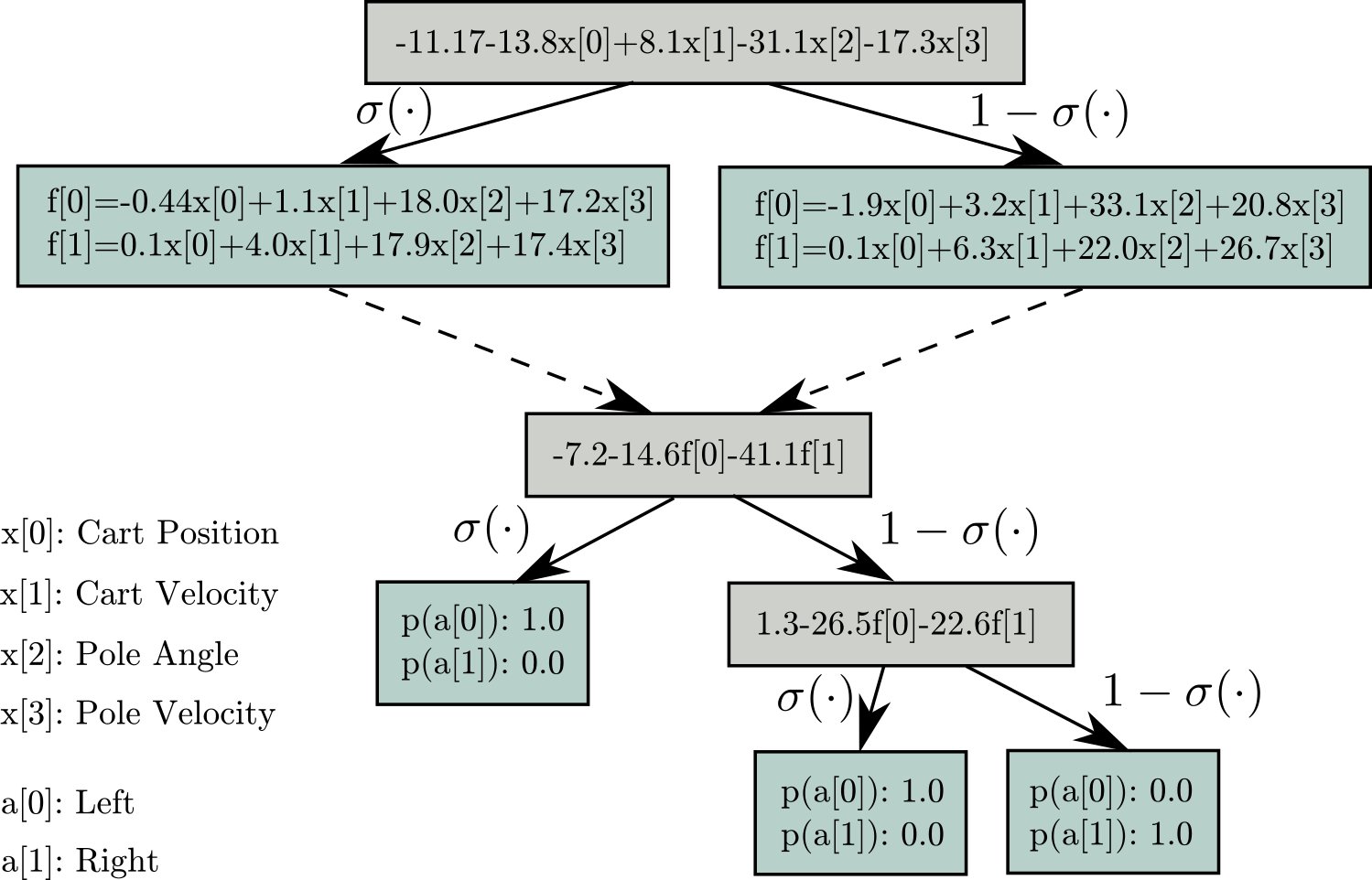}
    \caption{The learned CDT (before discretization) of depth 1+2 for \textit{CartPole-v1}. }
    \label{fig:cartpole_plot_cdt}
\end{figure}

\begin{figure}[htbp]
    \centering
        \centering
        \includegraphics[scale=0.24]{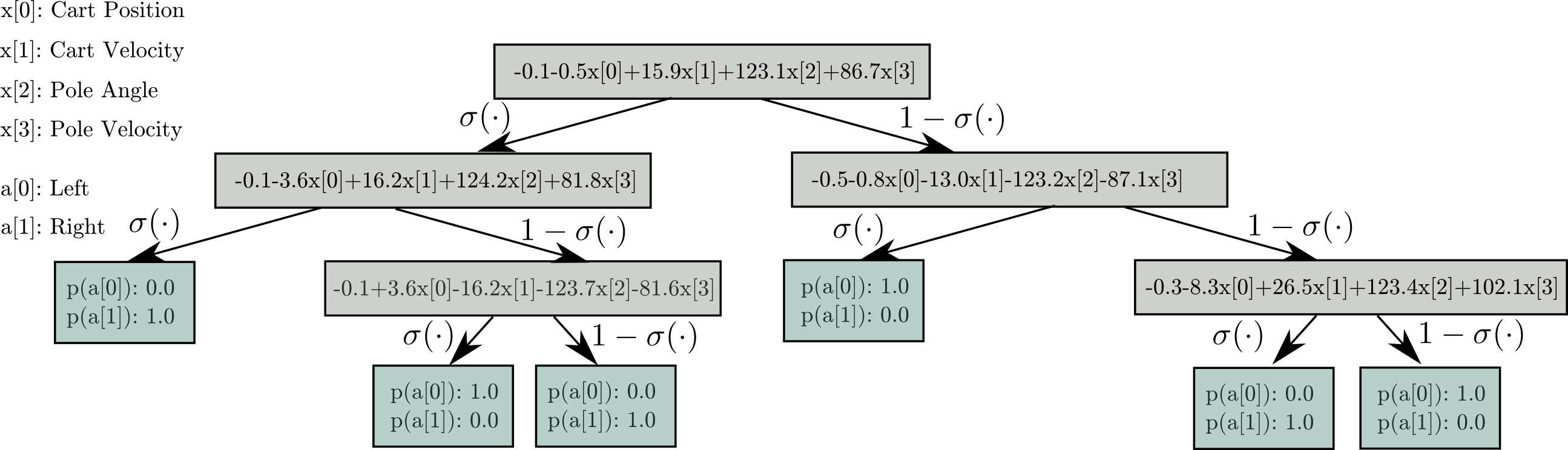}
    \caption{The learned SDT (before discretization) of depth 3 for \textit{CartPole-v1}. }
    \label{fig:cartpole_plot_sdt}
\end{figure}

\begin{figure}[htbp]
    \centering
        \centering
        \includegraphics[scale=0.24]{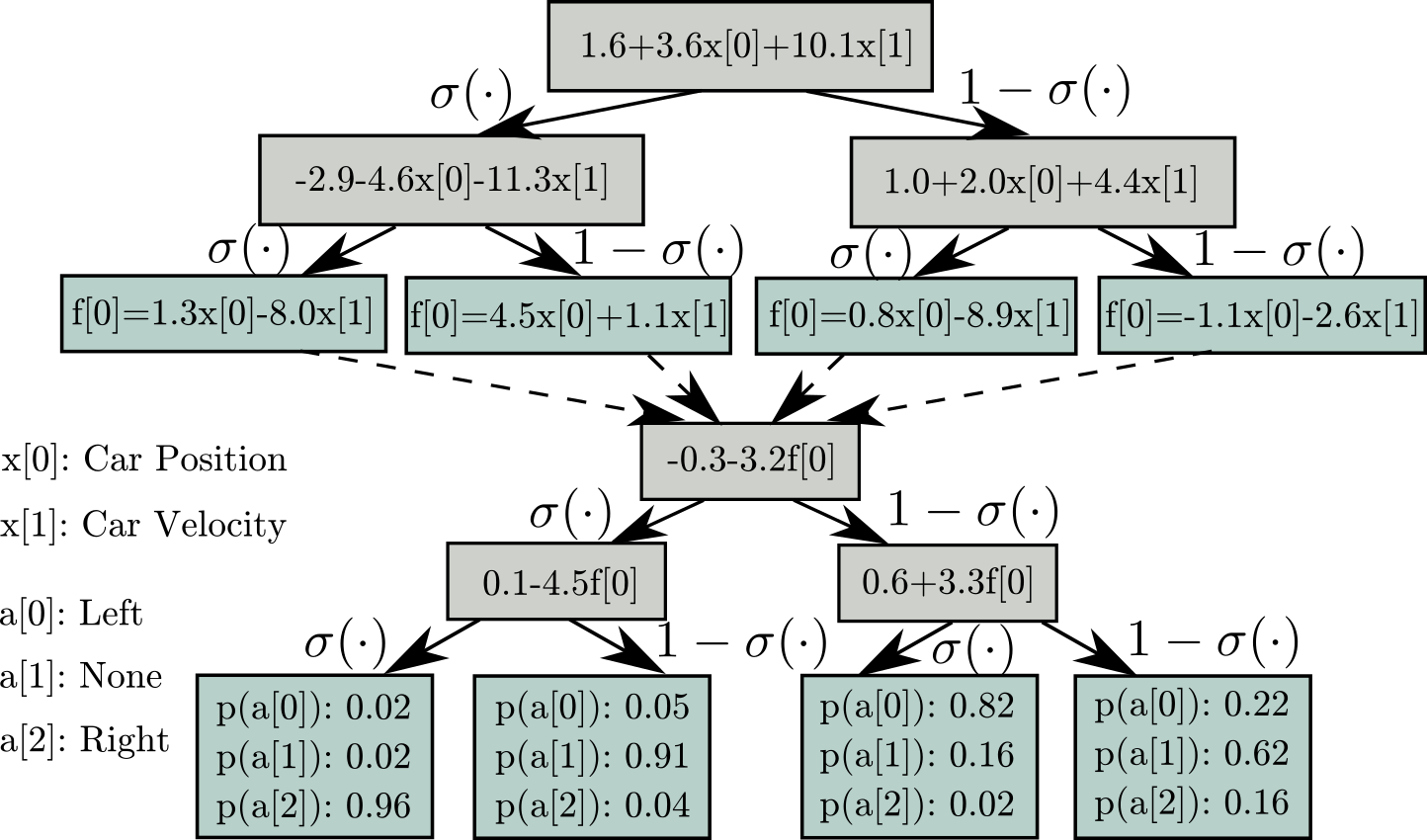}
    \caption{The learned CDT (before discretization) of depth 2+2 for \textit{MountainCar-v0}. }
    \label{fig:mountaincar_plot_cdt}
\end{figure}

% \begin{figure}[htbp]
%     \centering
%         \centering
%         \includegraphics[scale=0.18]{img/lunarlander_cdt_DiscretizedAll.png}
%     \caption{The learned CDT (after discretization) of depth 2+2 for \textit{LunarLander-v2}: two dimensions are reserved for weight vectors in both $\mathcal{F}$ and $\mathcal{D}$, as well as the intermediate features. }
%     \label{fig:lunarlander_plot_cdt}
% \end{figure}
\end{appendices}

\end{document}